\documentclass{article}
\usepackage{iclr2027_conference,times}
\usepackage{amsmath,amssymb,amsthm,booktabs}
\usepackage{algorithm,algorithmic,graphicx,hyperref,url,xcolor}
\hypersetup{hidelinks}

\newcommand{\method}{REG}

\title{
Robust Ensemble Guidance for
Scientific Inverse Problems}
\author{Zixiang Li$^{1}$, Wei Wang$^{1}$, Yunchao Wei$^{1}$, Yao Zhao$^{1}$, Yue Song$^{2}$\\
$^{1}$Beijing Jiaotong University\\
$^{2}$Tsinghua University}
\date{}

\iclrfinalcopy
\begin{document}
\maketitle
\fancyhead{}
\begin{abstract}
Ensemble guidance combines pretrained diffusion priors with black-box forward models to solve inverse problems without differentiating through the physical simulator. However, observation coordinates with large predictive spread or extreme residuals can dominate the ensemble correction, degrading reconstruction accuracy. We show that two simple modifications, weighting and clipping, substantially improve this correction. Our method, Robust Ensemble Guidance (REG), uses ensemble predictive spread to balance observation scales and adaptively clips standardized residuals to limit the influence of extreme discrepancies. Both operations reuse existing particles and forward predictions, requiring no additional denoiser or forward-model evaluations. Under a local linear Gaussian model, we derive conditions for reduced one-step estimation risk, bound the influence of individual observation coordinates, and characterize when these benefits persist with finite ensembles. Experiments on Navier–Stokes inversion, black-hole imaging, and acoustic full-waveform inversion demonstrate improved reconstruction over the underlying ensemble solver. In particular, REG \textbf{increases black-hole reconstruction PSNR by 6.2–8.2 dB} across three observation regimes and \textbf{reduces Navier–Stokes reconstruction error by 26.4\%} in a matched-budget comparison. These findings highlight the importance of observation heterogeneity and residual influence in designing reliable generative solvers for scientific inverse problems.
\end{abstract}

\begingroup
\raggedbottom
\setlength{\parskip}{3pt}
\setlength{\textfloatsep}{10pt plus 2pt minus 2pt}
\setlength{\floatsep}{8pt plus 2pt minus 2pt}
\setlength{\intextsep}{10pt plus 2pt minus 2pt}
\setlength{\abovedisplayskip}{5pt plus 1pt minus 1pt}
\setlength{\belowdisplayskip}{5pt plus 1pt minus 1pt}
\setlength{\belowdisplayshortskip}{3pt plus 1pt minus 1pt}
\section{Introduction}
\label{sec:introduction}

Inverse problems seek an unknown state $x^\star$ from an indirect, incomplete,
and noisy observation $y$ \citep{tarantola2005inverse}.  Since many states may explain the same
measurement, recovery requires a strong prior \citep{stuart2010inverse,bora2017compressed}. Pretrained diffusion models
\citep{ho2020ddpm,song2021score,karras2022edm} provide reusable priors for restoration,
sensing, and scientific inference \citep{song2022medical,jalal2021mri}.
The prior keeps reconstructions near the data distribution;
observation correction promotes measurement consistency \citep{kadkhodaie2021stochastic}.

Most observation corrections use a data-fidelity gradient. They propagate the
measurement error through a denoised prediction and the forward model
\citep{chung2023dps,chung2022mcg}.
Other methods use pseudoinverses \citep{song2023pigdm,wang2023ddnm}
or known operator structure \citep{kawar2022ddrm,choi2021ilvr}. These forms
of access are natural for many imaging operators, but they are not always
available. A PDE adjoint may be expensive to derive and store, and a discrete
or long-running simulator may provide unstable sensitivities. Such cases
require guidance from forward evaluations alone.

Ensemble guidance provides this capability by passing a set of generative
particles through the forward model \citep{zheng2024enkg}.
Empirical correlations between particle
states and predicted observations then transport the measurement error back to
the state space \citep{evensen2003enkf,iglesias2013ensemble,schillings2017analysis}.
This approach removes the need for a forward derivative, but it
introduces a different problem. The correction itself is estimated from
finitely many particles, and its behavior depends on how observation
coordinates and residual magnitudes enter the update.

Two effects can make this correction unstable. First, predicted observations
can vary on very different scales across coordinates. Without standardization,
a large-scale coordinate can dominate the interaction between a residual and
the ensemble predictions, amplifying any noise or estimation error in that
coordinate. A large spread need not imply weak information: the issue is how
unequal scales affect the combined update. Second, a quadratic data-fidelity
loss gives increasing influence to extreme residuals \citep{huber2009statistics,maronna2019robust}. A few corrupted
or poorly predicted measurements can therefore distort an otherwise useful
direction. A smaller step limits this error's size but not its direction.
These effects motivate balancing coordinate scales and limiting extreme
residuals before particle correction.
We introduce Robust Ensemble Guidance (\method{}), a derivative-free
correction designed for this finite-ensemble setting. REG uses the predictive
spread of the ensemble to standardize both the residuals and the
predicted-observation variations. This produces a diagonal observation
geometry that reduces the influence of high-spread coordinates. REG then
clips the standardized residuals at a shared, quantile-adaptive threshold,
limiting the influence of extreme discrepancies on the particle interaction.
Both operations reuse the particles and forward
predictions already needed for the correction. They add no denoiser call and
no forward-model evaluation. Uniform coordinate weights without clipping
recover the standard quadratic ensemble correction.

Our analysis explains when these modifications help. Under a local linear Gaussian
model, balancing predictive scales can improve conditioning and
reduce the error component constrained by the observations. We derive
conditions for lower one-step estimation risk and characterize when this
advantage persists under finite-ensemble estimation of the weights. Our
clipping analysis further establishes bounds on residual influence and
identifies conditions under which suppressing extreme residuals reduces
estimation error. Controlled experiments validate these mechanisms, while
experiments on Navier--Stokes inversion, black-hole imaging, and acoustic
full-waveform inversion demonstrate improved reconstruction over other methods.
These results establish \method{}'s effectiveness as a derivative-free
solver across diverse scientific inverse problems and observation regimes.

\section{Related Work}
\label{sec:related}

\paragraph{Diffusion priors for inverse problems.}
Diffusion models \citep{sohldickstein2015diffusion,song2019gradients,ho2020ddpm,song2021score,karras2022edm}
provide expressive priors that can be reused across observations. Guidance
methods combine this prior with an observation likelihood, building on
denoiser-based regularization \citep{romano2017red} and plug-and-play
reconstruction \citep{venkatakrishnan2013pnp}. A common strategy
forms a clean prediction at each diffusion step and differentiates its
measurement mismatch through both the denoiser and a known forward operator
\citep{chung2023dps,chung2022mcg}. Other formulations use variational optimization
\citep{mardani2024variational} or decouple prior updates from data consistency
\citep{zhu2023diffpir,li2024decoupled}. Known operator structure
\citep{kawar2022ddrm} and pseudoinverses \citep{song2023pigdm,wang2023ddnm}
also enable observation conditioning.
\method{} addresses the complementary derivative-free setting, using
generative particles and forward values.

\paragraph{Ensemble Kalman inversion.}
Ensemble Kalman methods infer state--observation relations from particles
\citep{iglesias2013ensemble,kovachki2019eki}. Their theory addresses convergence
\citep{schillings2017analysis,blomker2019wellposed} and mean-field limits
\citep{ding2021meanfield}. Iterative regularization \citep{iglesias2016regularizing}
and repeated data assimilation \citep{emerick2013smoother} extend this framework.
General-likelihood
extensions likewise modify the score transported through the empirical
cross-covariance \citep{duffield2022general}.  EnKG brings this statistical
transport to diffusion guidance using an identity observation geometry and a
quadratic residual \citep{zheng2024enkg}.
Sample covariance concentration quantifies finite-sample error
\citep{koltchinskii2017covariance}.
We study finite-ensemble observation scaling and residual influence at fixed
coordinates.

\paragraph{Robust estimation.}
Huber losses are quadratic near zero \citep{huber1964robust}; their bounded
scores limit residual influence \citep{hampel1974influence}. Related robust mean
\citep{catoni2012challenging} and regression estimators \citep{sun2020adaptive}
control heavy-tailed fluctuations.
Unlike oracle noise weights, inverse spread is observable but fallible.
Our analysis identifies when the geometry and bounded score
do or do not improve the conventional identity-quadratic update.

\section{Robust Ensemble Guidance}
\label{sec:method}

Robust Ensemble Guidance (\method{}) combines diffusion-prior prediction with
an ensemble correction. Paired state and observation variations transport
data discrepancy to the unknown state. Predictive-spread weighting balances
local observation scales, while quantile-adaptive clipping limits extreme
residuals. We derive these components and conditions for lower one-step risk.

\begin{figure}[t]
  \vspace{-3mm}
    \centering
    \includegraphics[width=\linewidth]{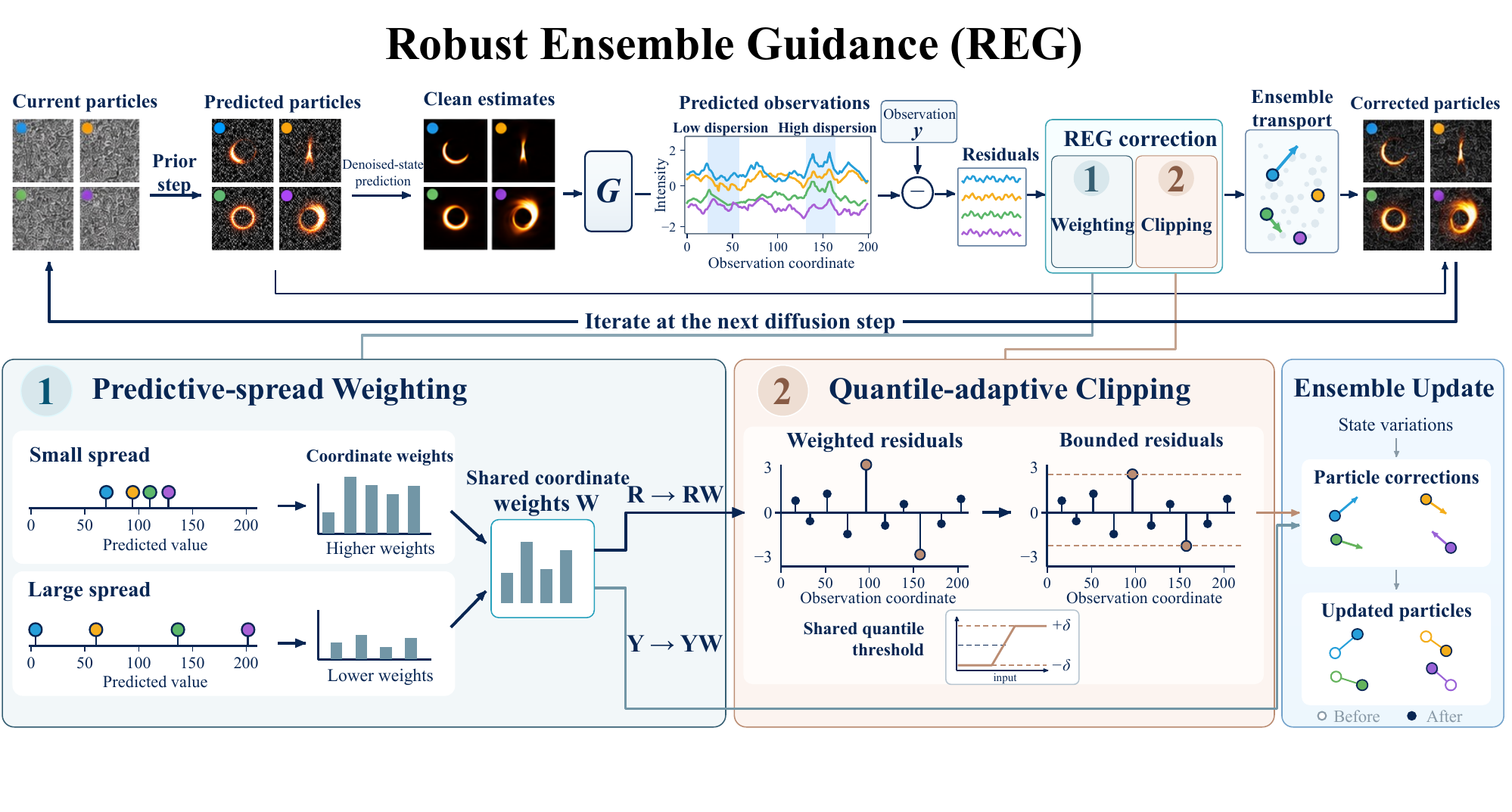}
    \vspace{-13mm}
    \caption{Overview of Robust Ensemble Guidance.  Forward predictions from
    a denoised particle ensemble determine coordinate weights and a shared
    clipping threshold.  The resulting bounded residuals are transported back
    to the particle space to form the next correction.}
    \label{fig:rpg_method_overview}
    \vspace{-1mm}
\end{figure}

\subsection{Ensemble guidance from forward predictions}
\label{sec:rpg_pc}

Let $G:\mathbb R^n\to\mathbb R^m$ denote the forward model, where $n$ and $m$
are the state and observation dimensions.  The unknown state $x^\star\in
\mathbb R^n$ produces the observation $y\in\mathbb R^m$ according to:
\begin{equation}
    y=G(x^\star)+\xi,
    \qquad \Gamma=\operatorname{Cov}(\xi),
    \label{eq:rpg_problem}
\end{equation}
where $\xi$ is zero-mean observation noise and $\Gamma\in\mathbb R^{m\times m}$
is its covariance.  The decreasing diffusion-time schedule
$\tau_T>\cdots>\tau_0=0$ comprises $T$ reverse steps.
At each step $t\in\{T,\ldots,1\}$, \method{} maintains an ensemble of $J$
particles $\{x_t^{(j)}\}_{j=1}^{J}$ indexed by $j$.  For example, one
Euler step of the probability-flow dynamics gives the diffusion prediction:
\begin{equation}
    x_{t-1}^{\mathrm{pred},(j)}=x_t^{(j)}-
    \dot\sigma(\tau_t)\sigma(\tau_t)
    s_\theta(x_t^{(j)},\tau_t)(\tau_{t-1}-\tau_t),
    \label{eq:rpg_prediction}
\end{equation}
Here $\sigma(\tau)$ is the diffusion noise schedule with time derivative
$\dot\sigma$. The score model $s_\theta$ has pretrained parameters $\theta$.
The superscript $\mathrm{pred}$ distinguishes the prediction from
the corrected particle $x_{t-1}^{(j)}$ below. The correction does not depend
on the numerical solver.

The denoising map $\phi_t$ operates at the predicted particle's noise level.
Correction quantities are indexed by reverse step $t$. The clean-state estimate
$\widehat x_t^{(j)}$, predicted observation $\widehat y_t^{(j)}$, and residual
$r_t^{(j)}$ are:
\begin{equation}
    \widehat x_t^{(j)}=\phi_t(x_{t-1}^{\mathrm{pred},(j)}),\qquad
    \widehat y_t^{(j)}=G(\widehat x_t^{(j)}),\qquad
    r_t^{(j)}=y-\widehat y_t^{(j)}.
    \label{eq:rpg_forward}
\end{equation}
$\bar x_{t-1}^{\mathrm{pred}}$ and $\bar y_t$ denote the predicted-particle and observation means.
The state-anomaly matrix $E_t\in\mathbb R^{J\times n}$,
observation-anomaly matrix $Y_t\in\mathbb R^{J\times m}$, and residual matrix
$R_t\in\mathbb R^{J\times m}$ have rows:
\begin{equation}
    (E_t)_{j,:}=(x_{t-1}^{\mathrm{pred},(j)}-
    \bar x_{t-1}^{\mathrm{pred}})^\top,\qquad
    (Y_t)_{j,:}=(\widehat y_t^{(j)}-\bar y_t)^\top,\qquad
    (R_t)_{j,:}=(r_t^{(j)})^\top.
    \label{eq:rpg_anomalies}
\end{equation}
The matrix $X_{t-1}^{\mathrm{pred}}\in\mathbb R^{J\times n}$ stacks the predicted
particles by row. The basic coupling matrix $\mathcal A_t^{(0)}\in\mathbb R^{J\times J}$
and the corrected particle matrix $X_{t-1}\in\mathbb R^{J\times n}$ are:
\begin{equation}
    \mathcal A_t^{(0)}=\frac1J R_tY_t^\top,\qquad
    X_{t-1}=X_{t-1}^{\mathrm{pred}}+\gamma_t \mathcal A_t^{(0)}E_t,
    \label{eq:rpg_basic_direction}
\end{equation}
where $\gamma_t>0$ is the correction step size.  The entry
$(\mathcal A_t^{(0)})_{j\ell}$ measures the alignment between particle $j$'s
residual and particle $\ell$'s observation anomaly. The corresponding state
anomaly of particle $\ell$ transfers this information to the state space.
The empirical state covariance $C_t^{xx}$ and state--observation
cross-covariance $C_t^{xy}$ are computed from the ensemble as:
\begin{equation}
    C_t^{xx}=\frac1J E_t^\top E_t,\qquad
    C_t^{xy}=\frac1J E_t^\top Y_t.
    \label{eq:rpg_covariances}
\end{equation}
For $G\circ\phi_t$ locally linear over the ensemble, its linearization
$H_t=\nabla(G\circ\phi_t)(\bar x_{t-1}^{\mathrm{pred}})\in
\mathbb R^{m\times n}$ gives:
\begin{equation}
    C_t^{xy}\approx C_t^{xx}H_t^\top,
    \label{eq:rpg_statistical_linearization}
\end{equation}
which supplies a covariance-preconditioned local direction without forming
$H_t$. Weighting and clipping control how observation coordinates and
residual magnitudes enter this direction.

The analysis considers a single local correction with fixed linearization
and covariance. A fixed step size $\alpha>0$ serves as the theoretical
counterpart of $\gamma_t$.
The local reference state $x_t^{\mathrm{ref}}$ is in the
particle space and satisfies $G(\phi_t(x_t^{\mathrm{ref}}))=G(x^\star)$.
This idealization isolates the
observation correction from denoising and linearization errors.
Appendix~\ref{app:theory} gives assumptions, proofs, and additional analysis.

\subsection{Reliability-weighted observation geometry}
\label{sec:rpg_weighting}
\label{sec:theory}

Identity observation weighting assigns the same coefficient to every
coordinate, even when their predictive scales differ substantially. The
resulting update can amplify a large-scale coordinate together with its
perturbations. A two-coordinate example makes the scale
instability explicit.  Consider a scalar local state with unit ensemble
variance and linearized observation vector $h=(1,L)^\top$, where $L>0$ is the
scale ratio between the two coordinates.  The two predictive standard
deviations are then proportional to $1$ and $L$.  Here $e$ denotes the scalar
mean-state error and $\zeta$ represents a perturbation in the second residual.
The identity-weighted correction $g_I$ is:
\begin{equation}
    g_I=(1+L^2)e+L\zeta.
    \label{eq:rpg_scale_example_identity}
\end{equation}
The corrected error is $e^+=e-\alpha g_I$. The iteration is
stable only when $0<\alpha<2/(1+L^2)$. The perturbation enters with gain
$L$. Thus a large spread ratio simultaneously narrows the stable step-size
range and amplifies finite-ensemble or measurement perturbations.

For observation coordinate $k\in\{1,\ldots,m\}$, empirical standard deviation
$v_{t,k}$ defines raw weight $\widetilde w_{t,k}$ and unit-mean normalized weight $w_{t,k}$:
\begin{equation}
    v_{t,k}=\operatorname{Std}_{j}[\widehat y_{t,k}^{(j)}],\qquad
    \widetilde w_{t,k}=\frac{1}{v_{t,k}+\varepsilon_w},\qquad
    w_{t,k}=\frac{\widetilde w_{t,k}}
    {m^{-1}\sum_{k'=1}^{m}\widetilde w_{t,k'}}.
    \label{eq:precision_weight}
\end{equation}
Here $\operatorname{Std}_j$ denotes the standard deviation across particles
and $k'$ indexes the sum. The positive floor $\varepsilon_w$ prevents
unbounded raw weights.
Unit-mean normalization separates relative coordinate importance from the
overall step size.  The diagonal weight matrix $W_t\in\mathbb R^{m\times m}$
and the weighted residual and observation-anomaly matrices
$\widetilde R_t,\widetilde Y_t\in\mathbb R^{J\times m}$ are defined by:
\begin{equation}
    W_t=\operatorname{diag}(w_{t,1},\ldots,w_{t,m}),\qquad
    \widetilde R_t=R_tW_t,\qquad
    \widetilde Y_t=Y_tW_t.
    \label{eq:rpg_weighted_terms}
\end{equation}
The weighted coupling $\widetilde R_t\widetilde Y_t^\top/J$ uses identical
weights for residuals and observation anomalies.

\paragraph{Why weighting helps.}
At a fixed reverse step, $C=C_t^{xx}$ and $H=H_t$ denote the fixed state
covariance and linearization. The prediction covariance is $\Sigma_y=HCH^\top$.
Applying the weight rule to the population spreads $v_k=\sqrt{(\Sigma_y)_{kk}}$
yields the population geometry $Q\in\mathbb R^{m\times m}$.
Its empirical counterpart is $\widehat Q=W_t^2$.
The diagonal variance matrix is $D=\operatorname{diag}(v_1^2,\ldots,v_m^2)$.
For positive spreads in the zero-floor limit, the normalization factor
$c=(m^{-1}\sum_k v_k^{-1})^{-1}$ gives:
\begin{equation}
    Q=c^2D^{-1}.
    \label{eq:rpg_diagonal_whitening}
\end{equation}
Thus weighting diagonally whitens observation space and removes arbitrary
coordinate scales in this limit. With a nonzero floor,
$Q_{kk}\propto(v_k+\varepsilon_w)^{-2}$.
For correlated observations, the magnitude of the standardized off-diagonal
correlations is measured by
$\rho=\|D^{-1/2}(\Sigma_y-D)D^{-1/2}\|_{\mathrm{op}}<1$.
The condition number $\kappa$ of the standardized covariance then satisfies:
\begin{equation}
    \kappa\!\left(D^{-1/2}\Sigma_yD^{-1/2}\right)
    \leq \frac{1+\rho}{1-\rho}.
    \label{eq:rpg_conditioning_bound}
\end{equation}
Hence weighting improves conditioning when marginal scale imbalance dominates
the standardized correlations.  In the noiseless diagonal case, the
update contracts covariance-whitened observable error for $0<\alpha c^2<2$.
The unobservable component remains unchanged.

Predictive spread is a local reliability proxy, so its
benefit under noise is conditional. Let
$e=\bar x_{t-1}^{\mathrm{pred}}-x_t^{\mathrm{ref}}$ denote the local mean-state error
and $e^+$ its value after correction. With $C,H,Q,\alpha$ fixed,
the one-step risk $\mathcal R(Q)$ averages squared error over observation noise:
\begin{equation}
    \mathcal R(Q):=\mathbb E\|e^+\|_2^2=
    \|(I_n-\alpha CH^\top QH)e\|_2^2
    +\alpha^2\operatorname{tr}
    \left(CH^\top Q\Gamma QHC\right),
    \label{eq:risk_short}
\end{equation}
The two terms are the
remaining state error and transferred observation noise.  Define the local
direction $d_Q=CH^\top QHe$ and transferred noise level
$\nu_Q=\operatorname{tr}(CH^\top Q\Gamma QHC)$. Setting $Q=I_m$ gives the identity-weighted
counterparts $d_I$ and $\nu_I$.
Risk expansion gives the exact criterion:
\begin{equation}
    2e^\top(d_Q-d_I)>
    \alpha\left(\|d_Q\|_2^2-\|d_I\|_2^2+\nu_Q-\nu_I\right).
    \label{eq:rpg_general_improvement_condition}
\end{equation}
The left side is the additional alignment with the state error. The right side
is the added magnitude and noise cost.  A positive alignment margin therefore
guarantees improvement for a sufficiently small step. For the empirical
geometry $\widehat Q=W_t^2$, define the population risk advantage
$\Delta_{\mathcal R}=\mathcal R(I_m)-\mathcal R(Q)>0$.  This advantage persists
whenever
$\|\widehat Q-Q\|_{\mathrm{op}}<\Delta_{\mathcal R}/L_{\mathcal R}$, where
$L_{\mathcal R}$ is the local sensitivity of risk to $Q$.  Under sub-Gaussian
predictions and nonvanishing spreads, the error is
$O(\sqrt{\log(m/\eta)/J})$ with probability at least $1-\eta$, where
$\eta\in(0,1)$ is the failure probability.

\subsection{Quantile-adaptive bounded residuals}
\label{sec:rpg_clipping}

Weighting does not bound residual influence. For a fixed quantile level
$q\in(0,1)$, \method{} computes the threshold $\delta_t$ from all weighted
residuals. The elementwise clipping function $\psi_{\delta_t}$ for $a\in\mathbb R$ is:
\begin{equation}
    \delta_t=\operatorname{Quantile}_{q}
    \left(\{|(\widetilde R_t)_{jk}|\}_{j,k}\right),\qquad
    \psi_{\delta_t}(a)=
    \operatorname{sign}(a)\min(|a|,\delta_t).
    \label{eq:weighted_huber}
\end{equation}
Clipping preserves signs and small entries, caps magnitudes at
$\delta_t$, and yields:
\begin{equation}
    \mathcal A_t=\frac1J\psi_{\delta_t}(\widetilde R_t)\widetilde Y_t^\top,
    \label{eq:rpg_robust_coupling}
\end{equation}
where $\mathcal A_t\in\mathbb R^{J\times J}$ couples residuals to particle anomalies.
A quantile adapts the threshold to each step's residual scale. Upper-tail
contamination below $1-q$ cannot make it unbounded.

\paragraph{Why clipping helps.}
For a fixed threshold $\delta\geq0$ and scalar residual $a$, clipping satisfies:
\begin{equation}
    a\,\psi_\delta(a)\geq \psi_\delta(a)^2\geq0.
    \label{eq:rpg_clipping_alignment}
\end{equation}
Clipping preserves residual signs while shrinking tails. With
$W_t,\widetilde Y_t,\delta_t$ fixed, changing the weighted residual entry
$(j,k)$ induces a coupling change $\Delta\mathcal A_t$.
Its Frobenius norm obeys:
\begin{equation}
    \|\Delta\mathcal A_t\|_F\leq
    \frac{2\delta_t}{J}\|(\widetilde Y_t)_{:,k}\|_2,
    \label{eq:influence_bound}
\end{equation}
Unlike step normalization, this bound limits an entry's influence, not just update length.

Clipping trades estimation variance for tail bias. Consider a random weighted
residual $u\in\mathbb R^m$ with clean distribution $P$. A fixed local map
$M\in\mathbb R^{n\times m}$ transfers observation scores to the state space.
For $J$ independent samples with fixed $M,\delta$,
clipping reduces sample-mean MSE if and only if:
\begin{equation}
    \frac1J\operatorname{tr}\!\left(
    M\bigl[\operatorname{Cov}_P(u)-
    \operatorname{Cov}_P(\psi_\delta(u))\bigr]M^\top\right)
    >\left\|M\mathbb E_P[u-\psi_\delta(u)]\right\|_2^2.
    \label{eq:rpg_clipping_improvement_condition}
\end{equation}
The left side is removed variance and the right side is squared clean-tail
bias.  Clipping is therefore favored when rare large residuals add more
variance than useful direction.

The trade-off remains bounded under arbitrary contamination.
The contaminated residual fraction is denoted by $\pi\in[0,1]$.
The clean clipping bias is
$b_\delta=\|\mathbb E_P[u-\psi_\delta(u)]\|_2$.  For $J$ residual samples
$\{u_j\}_{j=1}^J$, the clipped estimate
$\widehat g_\delta=MJ^{-1}\sum_j\psi_\delta(u_j)$ and the clean target
$g_P=M\mathbb E_Pu$ satisfy:
\begin{equation}
    \mathbb E\|\widehat g_\delta-g_P\|_2^2
    \leq \|M\|_{\mathrm{op}}^2
    \left[\left(b_\delta+2\pi\delta\sqrt m\right)^2
    +\frac{m\delta^2}{J}\right].
    \label{eq:rpg_clipping_mse_bound}
\end{equation}
The terms capture clean-tail bias, bounded contamination, and finite-ensemble
variance. Unlike the error for an unbounded linear residual,
this bound is independent of corrupted-residual magnitudes.

\subsection{Complete REG update}
\label{sec:rpg_complete}

Combining transport, weighting, and clipping yields:
\begin{equation}
    \mathcal A_t=\frac1J\psi_{\delta_t}(R_tW_t)(Y_tW_t)^\top,
    \qquad
    \gamma_t=\frac{\lambda_t}{\|\mathcal A_t\|_F+\varepsilon_\gamma},
    \qquad
    X_{t-1}=X_{t-1}^{\mathrm{pred}}+\gamma_t\mathcal A_tE_t,
    \label{eq:rpg_matrix_direction}
\end{equation}
where $\lambda_t>0$ is the prescribed coupling scale and
$\varepsilon_\gamma>0$ is the numerical floor used in step normalization.
The ensemble relation determines the state-space direction.
The weights $W_t$ balance observation scales while clipping bounds the
contribution of extreme residuals. The parameter $\lambda_t$ controls the normalized coupling; the
state-space displacement also depends on $E_t$.

\begin{algorithm}[t]
\caption{Robust Ensemble Guidance}
\label{alg:rpg}
\begin{algorithmic}[1]
\REQUIRE Forward model $G$, observation $y$, particles
$\{x_T^{(j)}\}_{j=1}^{J}$, denoisers $\{\phi_t\}_{t=1}^{T}$,
score model $s_\theta$, noise schedule $\sigma$, times $\{\tau_t\}_{t=0}^{T}$, floors
$\varepsilon_w,\varepsilon_\gamma$, quantile $q$, coupling scales
$\{\lambda_t\}_{t=1}^{T}$
\FOR{$t=T,T-1,\ldots,1$}
  \STATE Predict $x_{t-1}^{\mathrm{pred},(j)}$ by
  Equation~\eqref{eq:rpg_prediction}
  \STATE Evaluate $\widehat x_t^{(j)}=\phi_t(x_{t-1}^{\mathrm{pred},(j)})$ and
  $\widehat y_t^{(j)}=G(\widehat x_t^{(j)})$
  \STATE Form the state anomalies $E_t$, observation anomalies $Y_t$, and
  residuals $R_t$
  \STATE Compute $W_t$ by Equation~\eqref{eq:precision_weight}; set
  $\widetilde R_t\leftarrow R_tW_t$ and
  $\widetilde Y_t\leftarrow Y_tW_t$
  \STATE Compute $\delta_t$ by Equation~\eqref{eq:weighted_huber} and
  $\mathcal A_t\leftarrow\psi_{\delta_t}(\widetilde R_t)\widetilde Y_t^\top/J$
  \STATE $\gamma_t\leftarrow\lambda_t/(\|\mathcal A_t\|_F+\varepsilon_\gamma)$
  \STATE $X_{t-1}\leftarrow
  X_{t-1}^{\mathrm{pred}}+\gamma_t\mathcal A_tE_t$
\ENDFOR
\RETURN $\{x_0^{(j)}\}_{j=1}^{J}$
\end{algorithmic}
\end{algorithm}

\begingroup
\raggedbottom
\section{Experiments}
\label{sec:experiments}

We evaluate \method{} through controlled experiments and three scientific
inverse problems: Navier--Stokes inversion, black-hole imaging, and acoustic
full-waveform inversion. The controlled study isolates the effects of
weighting and clipping, while the scientific tasks assess reconstruction
quality across observation conditions. Particle-count comparisons and
component ablations further examine the contribution of each modification.

\paragraph{Baselines.}
We follow the comparison set in InverseBench~\citep{zheng2025inversebench}.
For \textbf{Navier--Stokes}, the baselines are EKI~\citep{iglesias2013ensemble},
DPS-fGSG and DPS-cGSG~\citep{zheng2024enkg,berahas2022gradient},
DPG~\citep{tang2023dpg}, SCG~\citep{huang2024scg}, and
EnKG~\citep{zheng2024enkg}. \textbf{Black-hole imaging} baselines are
SMILI~\citep{akiyama2019smili}, EHT-Imaging~\citep{chael2019ehtimaging},
DPS~\citep{chung2023dps}, LGD~\citep{song2023lgd},
RED-diff~\citep{mardani2024variational}, PnP-DM~\citep{wu2024pnpdm},
DAPS~\citep{zhang2024daps}, DiffPIR~\citep{zhu2023diffpir}, and
EnKG~\citep{zheng2024enkg}.  For \textbf{acoustic FWI}, we include Adam, DPS~\citep{chung2023dps},
LGD~\citep{song2023lgd}, DiffPIR~\citep{zhu2023diffpir},
DAPS~\citep{zhang2024daps}, PnP-DM~\citep{wu2024pnpdm},
RED-diff~\citep{mardani2024variational}, and EnKG~\citep{zheng2024enkg}.

\begin{figure}[thb]
    \vspace{-4pt}
    \centering
    \includegraphics[width=\linewidth]{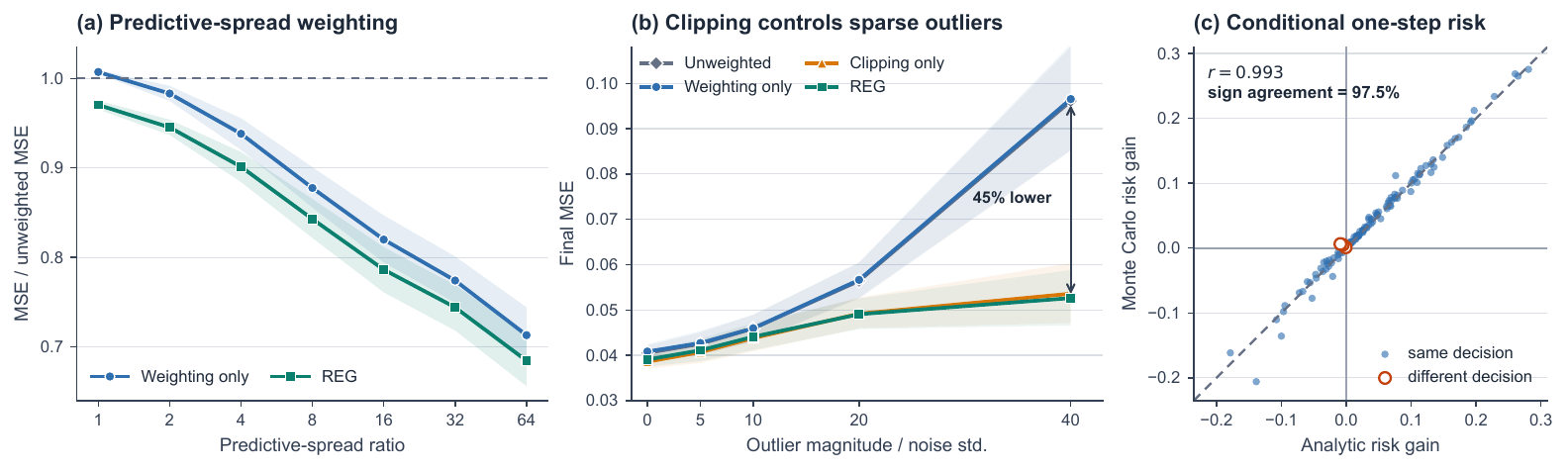}
    \vspace{-8mm}
    \caption{Controlled tests of weighting, clipping, and the conditional
    one-step risk criterion.}
    \label{fig:controlled_mechanisms}
    \vspace{-6pt}
\end{figure}

\subsection{Controlled validation of the REG mechanisms}
\label{sec:fixed_budget_experiments}

Finite-ensemble corrections can fail in two different ways. Unequal predictive
scales can cause some observation coordinates to dominate the update, while a
small number of extreme residuals can dominate an otherwise informative
update. The controlled study separates these effects and tests whether the two
parts of \method{} address the corresponding failure modes.

We use local linear Gaussian problems with $J=256$ particles in
Figure~\ref{fig:controlled_mechanisms}(a,b), with state dimension 16 and
observation dimension 64. Within each comparison, the methods share the
forward map, observation, initial ensemble, correction schedule, and random
seed. Panel (a) varies predictive-spread imbalance without contamination.
Panel (b) fixes equal predictive spread and varies the outlier magnitude
after corrupting 10\% of the residual coordinates. Both panels report final
ensemble-mean MSE with paired 95\% bootstrap confidence intervals.
Panel (c) separately compares analytic one-step risk gains with Monte Carlo
estimates on 120 independently generated problems.

\begin{table}[t]
\vspace{-4mm}
\centering
\caption{Navier--Stokes relative $\ell_2$ error ($\downarrow$).
Cells show mean above (s.d.); bold indicates the lowest mean in each column.}
\label{tab:navier}
\fontsize{9}{10}\selectfont
\setlength{\tabcolsep}{1.5pt}
\setlength{\aboverulesep}{1.2pt}
\setlength{\belowrulesep}{1.4pt}
\renewcommand{\arraystretch}{1.0}
\newcommand{\nsstat}[2]{\shortstack[c]{#1\\[-0.6pt]{\fontsize{7.5}{8}\selectfont(#2)}}}
\begin{tabular*}{\textwidth}{@{\extracolsep{\fill}}l*{9}{c}@{}}
\toprule
Subsampling
& \multicolumn{3}{c}{$\times2$}
& \multicolumn{3}{c}{$\times4$}
& \multicolumn{3}{c}{$\times8$} \\
\cmidrule(lr){2-4}\cmidrule(lr){5-7}\cmidrule(lr){8-10}
Method / $\sigma_y$
& $0$ & $1$ & $2$
& $0$ & $1$ & $2$
& $0$ & $1$ & $2$ \\
\midrule
EKI
& \nsstat{0.577}{0.138} & \nsstat{0.609}{0.119} & \nsstat{0.673}{0.107}
& \nsstat{0.579}{0.145} & \nsstat{0.669}{0.131} & \nsstat{0.805}{0.112}
& \nsstat{0.852}{0.167} & \nsstat{0.940}{0.115} & \nsstat{1.116}{0.090} \\
\midrule
DPS-fGSG
& \nsstat{1.687}{0.156} & \nsstat{1.612}{0.173} & \nsstat{1.454}{0.154}
& \nsstat{1.203}{0.122} & \nsstat{1.209}{0.116} & \nsstat{1.200}{0.100}
& \nsstat{1.246}{0.108} & \nsstat{1.221}{0.082} & \nsstat{1.260}{0.117} \\
DPS-cGSG
& \nsstat{2.203}{0.314} & \nsstat{2.117}{0.295} & \nsstat{1.746}{0.191}
& \nsstat{1.175}{0.079} & \nsstat{1.133}{0.095} & \nsstat{1.114}{0.144}
& \nsstat{1.186}{0.117} & \nsstat{1.204}{0.115} & \nsstat{1.218}{0.113} \\
DPG
& \nsstat{0.325}{0.188} & \nsstat{0.408}{0.173} & \nsstat{0.466}{0.171}
& \nsstat{0.322}{0.200} & \nsstat{0.361}{0.187} & \nsstat{\textbf{0.454}}{0.207}
& \nsstat{0.596}{0.301} & \nsstat{0.591}{0.262} & \nsstat{0.846}{0.251} \\
SCG
& \nsstat{0.908}{0.600} & \nsstat{0.928}{0.557} & \nsstat{0.966}{0.546}
& \nsstat{0.869}{0.513} & \nsstat{0.926}{0.546} & \nsstat{0.929}{0.505}
& \nsstat{1.260}{0.135} & \nsstat{1.284}{0.117} & \nsstat{1.347}{0.141} \\
EnKG
& \nsstat{0.116}{0.039} & \nsstat{\textbf{0.215}}{0.065} & \nsstat{0.341}{0.063}
& \nsstat{0.137}{0.115} & \nsstat{0.343}{0.130} & \nsstat{0.555}{0.103}
& \nsstat{0.317}{0.305} & \nsstat{\textbf{0.556}}{0.162} & \nsstat{0.763}{0.106} \\
\midrule
\textbf{\method{}}
& \nsstat{\textbf{0.100}}{0.031} & \nsstat{0.225}{0.103} & \nsstat{\textbf{0.334}}{0.080}
& \nsstat{\textbf{0.135}}{0.114} & \nsstat{\textbf{0.330}}{0.081} & \nsstat{0.506}{0.084}
& \nsstat{\textbf{0.272}}{0.261} & \nsstat{0.570}{0.185} & \nsstat{\textbf{0.757}}{0.109} \\
\bottomrule
\end{tabular*}
\vspace{-6pt}
\end{table}

Figure~\ref{fig:controlled_mechanisms} demonstrates the complementary roles
of weighting and clipping. Weighting offers little benefit with equal
predictive spreads, but improves reconstruction as scale imbalance grows (a).
Clipping controls the deterioration caused by increasingly large residual
outliers (b). These results show how coordinate scaling and residual influence
control address distinct sources of reconstruction error. The agreement
between analytic and Monte Carlo risk gains further supports the one-step
analysis (c).

\vspace{-4pt}

\subsection{Navier--Stokes inverse problem}
\label{sec:navier_stokes}

\paragraph{Problem setting.}
We recover initial vorticity from spatially subsampled, noisy observations of
two-dimensional incompressible Navier--Stokes flow. Nonlinear evolution and
loss of fine-scale information make recovery challenging, testing guidance
without differentiating through the PDE solver. We report relative $\ell_2$
error as mean and standard deviation across cases.

\vspace{-3pt}
\paragraph{Results.}
Table~\ref{tab:navier} compares relative $\ell_2$ error across nine
combinations of subsampling and noise. \method{} achieves the lowest mean
error in six settings and improves over EnKG in seven, including all
subsampling factors at $\sigma_y=2$. These gains cover both noiseless and
noisy observations, demonstrating improved recovery across varied
observation conditions. Also, figure~\ref{fig:navier_qualitative} complements the quantitative comparison: REG preserves the main vortex structures while reducing reconstruction errors relative to other methods.

\vspace{-3pt}
\paragraph{Particle-count comparison and component ablation.}
We next examine whether the complementary effects observed in the controlled study carry over to nonlinear flow inversion.
We compare \method{} with EnKG~\citep{zheng2024enkg} at
$\times2$ subsampling and $\sigma_y=0$. Both methods are evaluated on the same
cases with $J\in\{512,1024,2048\}$ particles. At $J=1024$,
\emph{Clipping only} and \emph{Weighting only} isolate
residual clipping and predictive-spread weighting, respectively.

\begin{table}[t]
\centering
\caption{Matched ablation ($\times2$, $\sigma_y=0$).
Relative $\ell_2$ error: mean $\pm$ s.d.; bold: lowest mean per comparison.
Changes are relative to EnKG at the same $J$.}
\label{tab:navier_ablation}
\fontsize{9}{10.5}\selectfont
\setlength{\tabcolsep}{1.5pt}
\renewcommand{\arraystretch}{1.12}
\begin{minipage}[t]{0.55\linewidth}
\centering
\textbf{(a) Particle-count comparison}\par\vspace{3pt}
\newcommand{\ablationrowstrut}{\rule[-5.488pt]{0pt}{15.68pt}}
\begin{tabular*}{\linewidth}{@{\extracolsep{\fill}}rccc@{}}
\toprule
 $J$ & EnKG & \method{} & Change \\
\midrule
\ablationrowstrut 512 & $0.2752\pm0.1088$ & $\mathbf{0.2690}\pm0.0890$ & $-2.3\%$ \\
\ablationrowstrut 1024 & $0.2029\pm0.1250$ & $\mathbf{0.1493}\pm0.0463$ & $-26.4\%$ \\
\ablationrowstrut 2048 & $0.1156\pm0.0385$ & $\mathbf{0.1000}\pm0.0314$ & $-13.4\%$ \\
\bottomrule
\end{tabular*}
\end{minipage}\hfill
\begin{minipage}[t]{0.42\linewidth}
\centering
\textbf{(b) Component ablation} ($J=1024$)\par\vspace{3pt}
\begin{tabular*}{\linewidth}{@{\extracolsep{\fill}}lcc@{}}
\toprule
Method & Relative $\ell_2\downarrow$ & Change \\
\midrule
EnKG & $0.2029\pm0.1250$ & -- \\
Clipping only & $0.1575\pm0.0576$ & $-22.4\%$ \\
Weighting only & $0.1670\pm0.0556$ & $-17.7\%$ \\
\textbf{\method{}} & $\mathbf{0.1493}\pm0.0463$ & $-26.4\%$ \\
\bottomrule
\end{tabular*}
\end{minipage}
\vspace{-6pt}
\end{table}

Table~\ref{tab:navier_ablation} shows lower mean error and lower variability
for \method{} than EnKG at every tested particle budget.  Both weighting and
clipping improve the correction individually, and their combination performs
best, supporting their complementary contribution beyond increasing ensemble
size. More particles improve both methods, but the relative advantage of
\method{} is not monotone in particle count.

\begin{figure}[t]
    \vspace{-3mm}
    \centering
    \includegraphics[width=\linewidth,trim=0 28bp 0 8bp,clip]{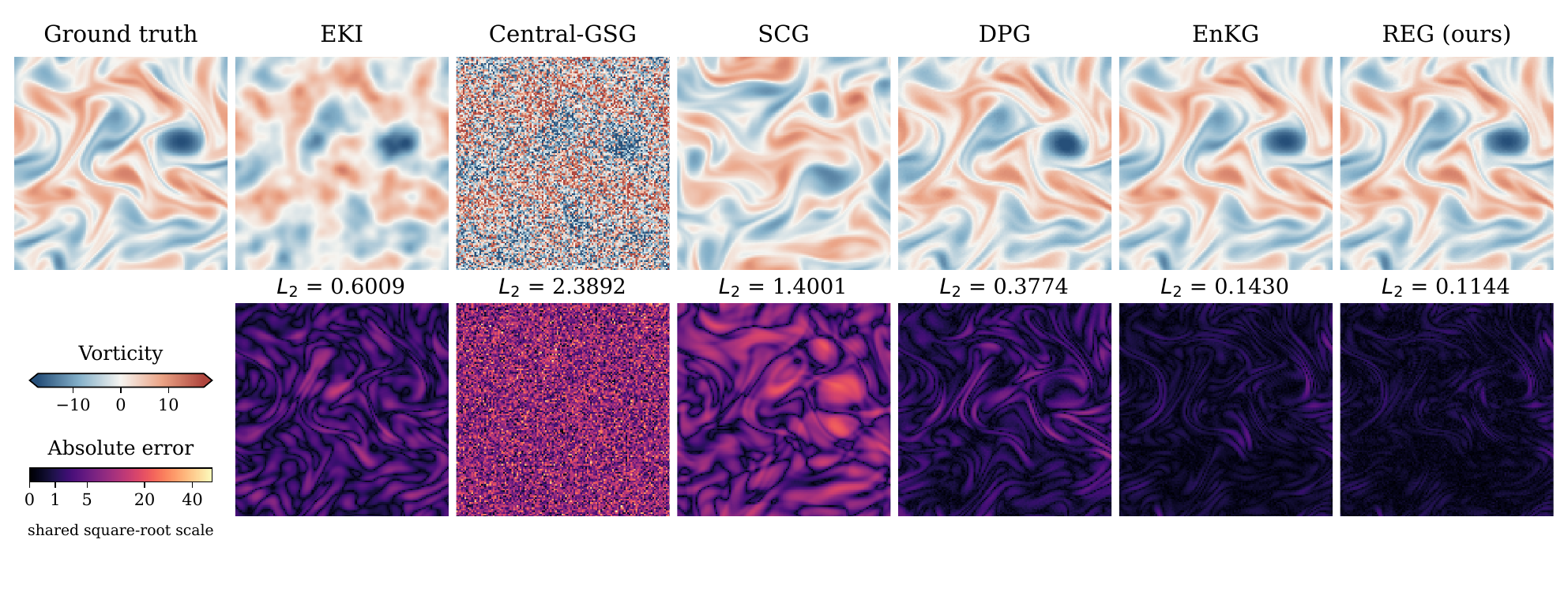}
    \vspace{-6mm}
    \caption{Navier--Stokes reconstructions (top) and absolute errors (bottom)
    for the same test case, with shared color scales across methods.}
    \label{fig:navier_qualitative}
    \vspace{-6pt}
\end{figure}

\subsection{Black-hole imaging}
\label{sec:blackhole}

\paragraph{Problem setting.}
We reconstruct black-hole images from sparse interferometric measurements
with atmospheric and thermal noise. Closure phases and log closure amplitudes
reduce atmospheric sensitivity but leave image details weakly constrained.
We follow InverseBench on 100 images at observation-time ratios of $3\%$,
$10\%$, and $100\%$, with $J=256$ particles for paired EnKG and \method{}.

\paragraph{Evaluation metrics.}
We report PSNR and its telescope-resolution blurred counterpart. Normalized
chi-squared values near one indicate agreement at the expected noise level.

\begin{table}[t]
\centering
\vspace{-2mm}
\caption{Black-hole imaging over 100 images. Cells show mean above (s.d.);
bold and underline mark the best and second-best PSNR and blurred PSNR (dB).}
\label{tab:blackhole}
\fontsize{8}{9}\selectfont
\setlength{\tabcolsep}{1pt}
\setlength{\aboverulesep}{1.2pt}
\setlength{\belowrulesep}{1.4pt}
\renewcommand{\arraystretch}{1.0}
\newcommand{\bhstat}[2]{\shortstack[c]{#1\\[-0.6pt]{\fontsize{7}{7.5}\selectfont(#2)}}}
\begin{tabular*}{\textwidth}{@{\extracolsep{\fill}}l*{12}{c}@{}}
\toprule
Obs. time
& \multicolumn{4}{c}{$\mathbf{3\%}$}
& \multicolumn{4}{c}{$\mathbf{10\%}$}
& \multicolumn{4}{c}{$\mathbf{100\%}$} \\
\cmidrule(lr){2-5}\cmidrule(lr){6-9}\cmidrule(lr){10-13}
Method
& PSNR$\uparrow$ & Blur$\uparrow$ & $\widetilde\chi^2_{cp}$ & $\widetilde\chi^2_{\log ca}$
& PSNR$\uparrow$ & Blur$\uparrow$ & $\widetilde\chi^2_{cp}$ & $\widetilde\chi^2_{\log ca}$
& PSNR$\uparrow$ & Blur$\uparrow$ & $\widetilde\chi^2_{cp}$ & $\widetilde\chi^2_{\log ca}$ \\
\midrule
SMILI
& \bhstat{18.51}{1.39} & \bhstat{23.08}{2.12} & \bhstat{1.478}{0.428} & \bhstat{4.348}{3.827}
& \bhstat{20.85}{2.90} & \bhstat{25.24}{3.86} & \bhstat{1.209}{0.169} & \bhstat{21.788}{12.491}
& \bhstat{22.67}{3.13} & \bhstat{27.79}{4.02} & \bhstat{1.878}{0.952} & \bhstat{17.612}{10.299} \\
EHT-Imaging
& \bhstat{21.72}{3.39} & \bhstat{25.66}{5.04} & \bhstat{1.507}{0.485} & \bhstat{1.695}{0.539}
& \bhstat{22.67}{3.46} & \bhstat{26.66}{3.93} & \bhstat{1.166}{0.156} & \bhstat{1.240}{0.205}
& \bhstat{24.28}{3.63} & \bhstat{28.57}{4.52} & \bhstat{1.251}{0.250} & \bhstat{1.259}{0.316} \\
\midrule
DPS
& \bhstat{24.20}{3.72} & \bhstat{\underline{30.83}}{5.58} & \bhstat{8.024}{24.336} & \bhstat{5.007}{5.750}
& \bhstat{24.36}{3.72} & \bhstat{30.79}{5.75} & \bhstat{13.052}{43.087} & \bhstat{6.614}{26.789}
& \bhstat{25.86}{3.90} & \bhstat{\underline{32.94}}{6.19} & \bhstat{8.759}{37.784} & \bhstat{5.456}{24.185} \\
LGD
& \bhstat{22.51}{3.76} & \bhstat{28.50}{5.49} & \bhstat{15.825}{16.838} & \bhstat{12.862}{12.663}
& \bhstat{22.08}{3.75} & \bhstat{27.48}{5.09} & \bhstat{10.775}{21.684} & \bhstat{13.375}{56.397}
& \bhstat{21.22}{3.64} & \bhstat{26.06}{4.98} & \bhstat{13.239}{17.231} & \bhstat{13.233}{39.107} \\
RED-diff
& \bhstat{20.74}{2.62} & \bhstat{26.10}{3.35} & \bhstat{6.713}{6.925} & \bhstat{9.128}{19.052}
& \bhstat{22.53}{3.02} & \bhstat{27.67}{4.53} & \bhstat{2.488}{2.925} & \bhstat{4.916}{13.221}
& \bhstat{23.77}{4.13} & \bhstat{29.13}{6.22} & \bhstat{1.853}{0.938} & \bhstat{2.050}{2.361} \\
PnP-DM
& \bhstat{\underline{24.25}}{3.45} & \bhstat{30.49}{4.93} & \bhstat{2.201}{1.352} & \bhstat{1.668}{0.551}
& \bhstat{\underline{24.57}}{3.47} & \bhstat{\underline{30.80}}{5.22} & \bhstat{1.433}{0.417} & \bhstat{1.336}{0.478}
& \bhstat{\underline{26.07}}{3.70} & \bhstat{32.88}{6.02} & \bhstat{1.311}{0.195} & \bhstat{1.199}{0.221} \\
DAPS
& \bhstat{23.54}{3.28} & \bhstat{29.48}{4.88} & \bhstat{3.647}{3.287} & \bhstat{2.329}{1.354}
& \bhstat{23.99}{3.56} & \bhstat{30.10}{5.13} & \bhstat{1.545}{0.705} & \bhstat{2.253}{9.903}
& \bhstat{25.60}{3.64} & \bhstat{32.78}{5.68} & \bhstat{1.300}{0.324} & \bhstat{1.229}{0.532} \\
DiffPIR
& \bhstat{24.12}{3.25} & \bhstat{30.45}{4.88} & \bhstat{14.085}{14.105} & \bhstat{10.545}{8.860}
& \bhstat{23.84}{3.59} & \bhstat{30.04}{5.03} & \bhstat{5.374}{3.733} & \bhstat{5.205}{5.556}
& \bhstat{25.01}{4.64} & \bhstat{31.86}{6.56} & \bhstat{3.271}{1.623} & \bhstat{2.970}{1.202} \\
EnKG
& \bhstat{20.77}{5.95} & \bhstat{25.58}{7.54} & \bhstat{14.019}{22.133} & \bhstat{23.559}{132.848}
& \bhstat{18.00}{9.18} & \bhstat{22.54}{11.88} & \bhstat{26.675}{41.330} & \bhstat{204.890}{675.821}
& \bhstat{21.79}{7.88} & \bhstat{27.48}{10.36} & \bhstat{25.173}{32.107} & \bhstat{29.759}{71.019} \\
\midrule
\textbf{\method{}}
& \bhstat{\textbf{26.97}}{3.79} & \bhstat{\textbf{34.08}}{5.56} & \bhstat{4.278}{7.335} & \bhstat{1.740}{1.786}
& \bhstat{\textbf{26.19}}{4.54} & \bhstat{\textbf{33.31}}{6.52} & \bhstat{7.198}{11.923} & \bhstat{2.908}{3.691}
& \bhstat{\textbf{28.16}}{4.58} & \bhstat{\textbf{36.84}}{6.96} & \bhstat{5.828}{9.397} & \bhstat{2.852}{2.726} \\
\bottomrule
\end{tabular*}
\end{table}

\paragraph{Results.}
Table~\ref{tab:blackhole} compares reconstruction quality and measurement
consistency at three observation-time ratios. REG achieves the highest PSNR and blurred PSNR in all three settings and improves measurement consistency over other methods. These results show improved image
recovery across observation durations, including the shortest observation
time.
The gains in both PSNR measures indicate improved fidelity at the original
image resolution and at the telescope-resolved scale.
Figure~\ref{fig:blackhole_qualitative} further shows improved preservation of the ring structure and brightness distribution.

\IfFileExists{figures/blackhole/blackhole_same_case_id1_all_methods.pdf}{%
\begin{figure}[t]
    \vspace{-4pt}
    \centering
    \includegraphics[width=\linewidth,trim=0 12bp 0 5bp,clip]{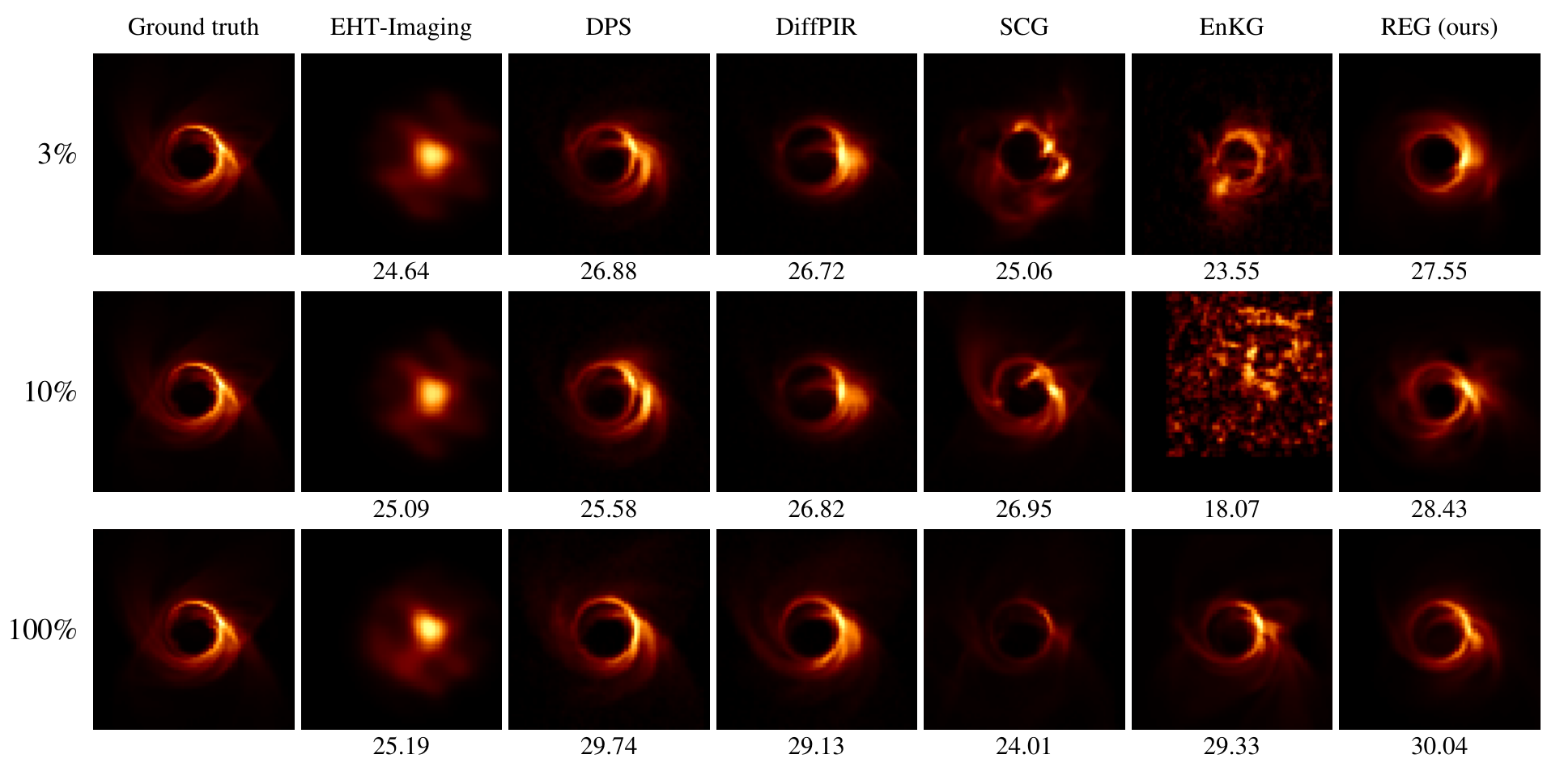}
    \vspace{-6pt}
    \caption{Selected black-hole reconstructions of one case at $3\%$, $10\%$, and
    $100\%$ observation time, with PSNR (dB). EnKG and \method{} share observations;
    other baselines retain native realizations.}
    \label{fig:blackhole_qualitative}
    \vspace{-6pt}
\end{figure}
}{}

\vspace{-3mm}
\subsection{Acoustic full-waveform inversion}
\label{sec:fwi}

\paragraph{Problem setting.}
We recover subsurface velocity maps from seismic measurements using the
InverseBench acoustic wave model. Sparse recordings and initialization
sensitivity make recovery challenging. Paired EnKG and \method{} use $J=64$
particles on ten official test cases.

\paragraph{Evaluation metrics.}
We report relative $\ell_2$, PSNR, SSIM, and seismic data misfit. For paired
EnKG and \method{}, image metrics are computed on each ensemble-mean
reconstruction, then averaged over test cases. Published baselines retain the
official benchmark evaluation.

\begin{table}[t]
\centering
\caption{Acoustic FWI results. Mean (s.d.).
Bold marks the best mean in each column.}
\label{tab:fwi}
\normalsize
\setlength{\tabcolsep}{3pt}
\renewcommand{\arraystretch}{1.03}
\begin{tabular*}{\linewidth}{@{\extracolsep{\fill}}lrrrr@{}}
\toprule
Method & Relative $\ell_2\downarrow$ & PSNR (dB) $\uparrow$ & SSIM $\uparrow$ & Data misfit $\downarrow$ \\
\midrule

Adam  & 0.333 (0.086) & 9.968 (2.083) & 0.305 (0.120) & 115.14 (52.10) \\
DPS  & 0.250 (0.154) & 14.111 (6.820) & 0.491 (0.161) & 155.08 (92.17) \\
LGD  & 0.244 (0.024) & 12.288 (0.889) & 0.341 (0.047) & 258.47 (26.40) \\
DiffPIR  & 0.204 (0.129) & \textbf{16.113} (6.962) & 0.554 (0.191)
 & \textbf{88.53} (56.91) \\
DAPS  & 0.201 (0.103) & 14.914 (4.184) & 0.321 (0.067) & 111.13 (71.33) \\
PnP-DM  & 0.259 (0.075) & 11.983 (2.269) & 0.431 (0.073) & 308.84 (26.34) \\
RED-diff  & 0.319 (0.102) & 10.372 (2.650) & 0.280 (0.108) & 94.67 (41.33) \\
EnKG  & 0.170 (0.034) & 15.437 (1.308) & 0.565 (0.065) & 269.64 (51.02) \\
\midrule
\textbf{\method{}}  & \textbf{0.160} (0.027) & 15.903 (1.084) & \textbf{0.574} (0.064) & 268.85 (49.51) \\
\bottomrule
\end{tabular*}
\end{table}

\paragraph{Results.}
Table~\ref{tab:fwi} compares velocity-map reconstruction and seismic data
misfit over ten test cases. \method{} achieves the lowest mean relative error
and highest mean SSIM among the compared methods, while improving relative
error, PSNR, and SSIM over EnKG. Relative error and SSIM assess complementary aspects of recovery. Their joint
improvement reflects more accurate velocity values together with better
structural agreement.
These results demonstrate improved
velocity-map recovery with an ensemble of only $64$ particles. Additional reconstructions and absolute-error maps are provided in
Appendix~\ref{app:qualitative_fwi}.
\endgroup

\section{Conclusion}
\label{sec:conclusion}

We presented REG for generative inverse problems.
REG combines predictive-spread weighting with adaptive residual clipping to
balance observation scales and limit the influence of extreme discrepancies.
Both reuse the ensemble and its forward predictions,
requiring neither forward derivatives nor additional model evaluations.
Under a local linear Gaussian model, our analysis identifies conditions for
reduced one-step estimation risk, bounds coordinate influence, and establishes
when improvement persists with a finite ensemble. Experiments support its
effectiveness across varied forward operators, observation
regimes, and problem scales.

\par
\endgroup
\clearpage
\subsection*{AI use statement}

Generative AI tools were used during manuscript preparation to assist with language polishing, grammatical correction, and LaTeX formatting. All technical content, methodological design, experimental implementation, result analysis, and conclusions were determined and verified by the authors.

\subsection*{Ethics statement}

This work studies scientific inverse problems using controlled simulations
and existing benchmarks for fluid flow, black-hole imaging, and seismic
inversion. The reported experiments do not involve recruiting human
participants or collecting personal data.

\subsection*{Reproducibility statement}

Section~\ref{sec:method} defines the particle prediction, weighting, clipping,
and correction steps, as summarized in
Algorithm~\ref{alg:rpg}. Appendix~\ref{app:theory} provides the assumptions
and proofs for the local analysis, distinguishing fixed-step risk calculations
from normalized algorithmic updates.
Section~\ref{sec:fixed_budget_experiments} describes the controlled tests and
their paired evaluation. Sections~\ref{sec:navier_stokes}--\ref{sec:fwi}
identify the scientific benchmarks, observation regimes, evaluation metrics,
and comparison protocols. Section~\ref{sec:experiments} cites benchmark and
baseline sources and distinguishes paired evaluations from published baseline results.

\bibliography{references}
\bibliographystyle{iclr2027_conference}
\appendix
\clearpage
\newpage
\section*{Supplementary material}

This supplement uses the notation of the main paper.
Appendix~\ref{app:theory} provides assumptions and proofs;
Appendix~\ref{app:experimental_details} describes experimental protocols
and evaluation conventions; Appendix~\ref{app:qualitative} presents additional
reconstructions.

\section{Assumptions and proofs for the local analysis}
\label{app:theory}

This appendix specifies the scope of the statements in
Section~\ref{sec:method}. The weighting results concern a frozen local linear
model. The clipping results first hold with the transport and threshold fixed.
We then discuss an estimated threshold.

\subsection{Local reference, linearization, and step normalization}
\label{app:local_model}

Fix a reverse step and write $F=G\circ\phi_t$ for the denoised forward map,
$C=C_t^{xx}$ for state covariance, and $H=H_t$ for its local linearization.
We omit $t$ on $E,Y,R,W,\widetilde Y,\delta,\mathcal A$ and $C^{xy}$ when
referring to the corresponding quantities at this fixed step.
The analysis uses the affine model
$F(x)=F(x_t^{\mathrm{ref}})+H(x-x_t^{\mathrm{ref}})$ on the region of
interest and assumes $F(x_t^{\mathrm{ref}})=G(x^\star)$. Thus the local target
is a state in the particle coordinates whose denoised prediction has the
correct observation. This assumption does not identify a noisy particle with
the clean state. Centered rows give $Y=EH^\top$, hence:
\begin{equation}
 C^{xy}=J^{-1}E^\top Y=J^{-1}E^\top EH^\top=CH^\top.
\end{equation}
For nonlinear $F$, remainder matrix $B_{\mathrm{nl}}$ gives
$Y=EH^\top+B_{\mathrm{nl}}$ and
$C^{xy}=CH^\top+J^{-1}E^\top B_{\mathrm{nl}}$.

Let $e=\bar x_{t-1}^{\mathrm{pred}}-x_t^{\mathrm{ref}}$. If the mean forward
prediction differs from $G(x^\star)+He$ by a deterministic mismatch
$b\in\mathbb R^m$, the mean residual is $-He-b+\xi$.
Here $\alpha$ is the fixed theoretical counterpart of $\gamma_t$,
$Q$ is a deterministic observation geometry, and $e^+$ is the error after
correction. The update gives
\begin{equation}
 e^+=(I_n-\alpha CH^\top QH)e-\alpha CH^\top Qb
       +\alpha CH^\top Q\xi.
 \label{eq:app_mismatch}
\end{equation}
The main-text risk sets $b=0$ and fixes $C,H,Q,\alpha$, averaging over
zero-mean noise of covariance $\Gamma$. A nonzero
$b$ contributes to the deterministic error term and can change the ranking
of two geometries. Dependence of these quantities on the same noise requires
further analysis.

The implemented normalization in Equation~\eqref{eq:rpg_matrix_direction}
obeys
\begin{equation}
 \|\gamma_t \mathcal A_t E_t\|_F
 \leq \lambda_t\frac{\|\mathcal A_t\|_F}{\|\mathcal A_t\|_F+\varepsilon_\gamma}
                  \|E_t\|_{\mathrm{op}}
 \leq \lambda_t\|E_t\|_{\mathrm{op}}.
\end{equation}
It bounds the displacement relative to the ensemble spread, rather than
setting its length to $\lambda_t$. The risk comparison uses a common fixed
$\alpha$.

\subsection{Scale balancing and conditioning}
\label{app:whitening}

Let $\Sigma_y=HCH^\top$ be prediction covariance, $v_k^2=(\Sigma_y)_{kk}$,
and $D=\operatorname{diag}(v_1^2,\ldots,v_m^2)$, with $D_{kk}>0$.
The population geometry $Q$ uses these population spreads; its empirical
counterpart is $\widehat Q=W_t^2$. In the idealized zero-floor limit, define
$c=(m^{-1}\sum_kD_{kk}^{-1/2})^{-1}$. Equation~\eqref{eq:precision_weight}
applied to population spreads gives $Q=c^2D^{-1}$. Thus
\begin{equation}
 Q^{1/2}\Sigma_yQ^{1/2}=c^2D^{-1/2}\Sigma_yD^{-1/2}.
\end{equation}
The diagonal entries are equal, while off-diagonal correlations remain.
This is marginal standardization, not a claim that diagonal weights remove
all correlations. With $\varepsilon_w>0$, the exact weights are proportional
to $(\sqrt{D_{kk}}+\varepsilon_w)^{-1}$; Equation~\eqref{eq:rpg_diagonal_whitening}
is the zero-floor limit.

Write $D^{-1/2}\Sigma_yD^{-1/2}=I_m+K$, where $K$ is symmetric and
$\|K\|_{\mathrm{op}}=\rho<1$. For unit vector $z$,
\begin{equation}
 1-\rho\leq z^\top(I_m+K)z\leq1+\rho.
\end{equation}
The smallest and largest eigenvalues therefore lie in this interval,
proving Equation~\eqref{eq:rpg_conditioning_bound}. A sufficient condition
for a strict improvement over the unweighted covariance is
$\kappa(\Sigma_y)>(1+\rho)/(1-\rho)$.

For positive coordinate rescaling $S=\operatorname{diag}(s_1,\ldots,s_m)$,
let $R'=RS$, $Y'=YS$. In the zero-floor limit $W'=aS^{-1}W$ for some common
$a>0$, so $R'W'=aRW$, $Y'W'=aYW$, and $\delta'=a\delta$. Homogeneity of
clipping gives $\mathcal A'=a^2\mathcal A$, where primes denote the rescaled
quantities. Thus $\mathcal A/\|\mathcal A\|_F$ is
invariant when $\mathcal A\ne0$ and the normalization floor is zero.

\subsection{Contraction of observable error}
\label{app:contraction}

Assume $C$ is positive definite and $\Sigma_y=HCH^\top=D$ is diagonal and
positive definite, with $Q=c^2D^{-1}$ as above. Define $B=HC^{1/2}$ and the covariance-whitened
error $z=C^{-1/2}e$. Since $BB^\top=D$, the orthogonal projector onto the row space of $B$ is:
\begin{equation}
 \Pi=B^\top(BB^\top)^{-1}B
\end{equation}
It satisfies $\Pi^\top=\Pi$ and
$\Pi^2=\Pi$. Writing $z^+=C^{-1/2}e^+$ for the whitened error after correction,
the noiseless update and its squared error satisfy:
\begin{equation}
 z^+=(I_n-\alpha c^2\Pi)z,
 \qquad
 \|z^+\|_2^2=(1-\alpha c^2)^2\|\Pi z\|_2^2+\|(I_n-\Pi)z\|_2^2.
\end{equation}
For $0<\alpha c^2<2$, the observable component contracts strictly unless it is
already zero; the null-space component is unchanged. We retain $c^2$
explicitly so $\alpha$ has the same meaning as in the risk analysis. For singular
$C$, the same argument applies on its range, provided $H$ has full row rank
on that range and the error lies in it.

\subsection{One-step risk and finite-ensemble weight estimation}
\label{app:risk}

Set $b=0$ in Equation~\eqref{eq:app_mismatch}. The mixed term vanishes because
$\mathbb E\xi=0$, and the identity
$\mathbb E\|CH^\top Q\xi\|_2^2=
\operatorname{tr}(CH^\top Q\Gamma QHC)$ gives
Equation~\eqref{eq:risk_short}. Only the first two noise moments are needed;
Gaussian noise is a sufficient special case. Expanding the deterministic
square yields
\begin{equation}
 \mathcal R(Q)=\|e\|_2^2-2\alpha e^\top d_Q
                  +\alpha^2(\|d_Q\|_2^2+\nu_Q).
\end{equation}
Subtracting $\mathcal R(I_m)$ proves
Equation~\eqref{eq:rpg_general_improvement_condition}. In particular, if
$a=e^\top(d_Q-d_I)>0$ and
$b_1=\|d_Q\|_2^2-\|d_I\|_2^2+\nu_Q-\nu_I$, improvement holds for
$0<\alpha<2a/b_1$ when $b_1>0$, and for every positive $\alpha$ when
$b_1\leq0$ in this frozen model.

For completeness, the local sensitivity in the main text can be made
explicit. Set $U=CH^\top$ for the frozen transport, and let $B_Q>0$ bound
the geometries: $\|Q\|_{\mathrm{op}},\|\widehat Q\|_{\mathrm{op}}\leq B_Q$. The difference
of squared norms and the Frobenius representation of the noise term imply
\begin{align}
 |\mathcal R(\widehat Q)-\mathcal R(Q)|
 &\leq L_{\mathcal R}\|\widehat Q-Q\|_{\mathrm{op}},\\
 L_{\mathcal R}
 &=2\alpha\|U\|_{\mathrm{op}}\|He\|_2
       (\|e\|_2+\alpha\|U\|_{\mathrm{op}}B_Q\|He\|_2)
       +2\alpha^2\|U\|_{\mathrm{op}}^2B_Q\operatorname{tr}(\Gamma).
\end{align}
If $L_{\mathcal R}>0$, a population advantage $\Delta_{\mathcal R}>0$
therefore persists when
$\|\widehat Q-Q\|_{\mathrm{op}}<\Delta_{\mathcal R}/L_{\mathcal R}$.
If $L_{\mathcal R}=0$, risk is constant on the neighborhood considered.

The stated rate follows under independent, identically distributed calibration
predictions with uniformly bounded sub-Gaussian norms and marginal standard
deviations $v_k$ in $[v_{\min},v_{\max}]$, where $v_{\min}>0$.
Write $\widehat v_k=v_{t,k}$ for their empirical estimates. Concentration of
the empirical first and second moments, followed by a union bound over
coordinates, gives
\begin{equation}
 \max_k|\widehat v_k^2-v_k^2|
 \leq C_0\left(\sqrt{\frac{\log(2m/\eta)}{J}}
                       +\frac{\log(2m/\eta)}{J}\right)
\end{equation}
with probability at least $1-\eta$. Here $C_0$ depends on the uniform
sub-Gaussian bound. This follows by applying Bernstein's inequality to the
squared centered coordinates and a sub-Gaussian bound to each sample mean.
On the event that empirical variances remain above $v_{\min}^2/2$, square
root, inverse scale, unit-mean normalization, and squaring are Lipschitz on
this bounded interval. The diagonal operator norm is the largest coordinate
error, so for $J$ large enough relative to
$\log(2m/\eta)$,
\begin{equation}
 \|\widehat Q-Q\|_{\mathrm{op}}
 =O\!\left(\sqrt{\frac{\log(2m/\eta)}{J}}\right).
\end{equation}
The constants depend on the spread bounds and the chosen floor. Population
and empirical weights must use the same floor; otherwise its deterministic
approximation error must also be included. This argument controls weight
estimation with $C,H,e$ fixed.

\subsection{Clipping influence and adaptive thresholds}
\label{app:influence}

For scalar residual $a$ with $|a|\leq\delta$,
$a\psi_\delta(a)=\psi_\delta(a)^2=a^2$; otherwise
$a\psi_\delta(a)=|a|\delta\geq\delta^2$. This proves
Equation~\eqref{eq:rpg_clipping_alignment}. Fix $W$, $\widetilde Y$, and
$\delta$, and change only weighted residual entry $(j,k)$ from $a$ to $a'$.
If $\mathbf e_j$ is the $j$th coordinate unit vector in $\mathbb R^J$, then
\begin{equation}
 \Delta\mathcal A=\frac{\psi_\delta(a')-\psi_\delta(a)}{J}
                  \mathbf e_j(\widetilde Y_{:,k})^\top.
\end{equation}
Since the scalar difference is at most $2\delta$ in magnitude, the norm of
this rank-one matrix proves Equation~\eqref{eq:influence_bound}.

Recomputing the threshold also changes other clipped entries. Let
$\widetilde R'$ differ from $\widetilde R=RW$ in one entry, with recomputed
threshold $\delta'$ and coupling $\mathcal A'$.
Adding and subtracting $\psi_\delta(\widetilde R')$ yields
\begin{equation}
 \|\mathcal A'-\mathcal A\|_F\leq\frac{2\delta}{J}\|\widetilde Y_{:,k}\|_2
       +\sqrt{\frac{m}{J}}\,|\delta'-\delta|
                          \|\widetilde Y\|_{\mathrm{op}},
\end{equation}
because $|\psi_{\delta'}(a)-\psi_\delta(a)|\leq|\delta'-\delta|$.

At the population level, under an upper-tail contamination fraction $\pi$,
a quantile level $q<1-\pi$ remains below a finite clean-distribution quantile,
provided that quantile is finite. For $N=Jm$ finite entries and $r$ arbitrary
replacements, the analogous condition is that the order-statistic indices
used by the empirical quantile lie at or below $N-r$. For example, for linear
interpolation at index $1+(N-1)q$, it suffices that
$\lceil1+(N-1)q\rceil\leq N-r$. This states the finite-sample rounding
condition behind the contamination-fraction interpretation. A shared measurement
can affect multiple entries, so the fraction concerns threshold entries.

\subsection{Variance, tail bias, and contamination risk}
\label{app:clipping_risk}

For Equation~\eqref{eq:rpg_clipping_improvement_condition}, assume
$u_1,\ldots,u_J$ are independent samples from the clean distribution $P$
with finite second moment, and fix $M$ and $\delta$ independently of these
samples. Write $\mu=\mathbb E_Pu$ and $\mu_\delta=\mathbb E_P\psi_\delta(u)$.
The unclipped mean is unbiased for $M\mu$; clipping adds bias
$M(\mu_\delta-\mu)$. Therefore
\begin{align}
 \mathbb E\left\|M J^{-1}\sum_j u_j-M\mu\right\|_2^2
 &=J^{-1}\operatorname{tr}(M\operatorname{Cov}_P(u)M^\top),\\
 \mathbb E\left\|M J^{-1}\sum_j\psi_\delta(u_j)-M\mu\right\|_2^2
 &=\|M(\mu_\delta-\mu)\|_2^2
   +J^{-1}\operatorname{tr}(M\operatorname{Cov}_P(\psi_\delta(u))M^\top).
\end{align}
Subtracting proves the exact criterion in the main text, with both
covariances taken under $P$.

For a scalar residual $a$, clipping is 1-Lipschitz, so an independent copy $a'$
gives
$\operatorname{Var}(\psi_\delta(a))=\tfrac12\mathbb E(\psi_\delta(a)-
\psi_\delta(a'))^2\leq\tfrac12\mathbb E(a-a')^2=\operatorname{Var}(a)$.
This reduces each marginal variance, but does not imply a covariance ordering
after an arbitrary map $M$ when coordinates are correlated. If coordinates
are independent, the covariance matrices are diagonal and the transported
variance cannot increase. If they are also symmetric about zero, the clipping
bias vanishes and the risk cannot increase, with strict improvement when a
direction used by $M$ has strictly reduced variance. For general residuals,
the explicit variance--bias criterion is required.

For Equation~\eqref{eq:rpg_clipping_mse_bound}, assume instead independent
samples from $(1-\pi)P+\pi P_{\mathrm{bad}}$, where $P$ has finite first
moment and $P_{\mathrm{bad}}$ is arbitrary. Clipping satisfies
$\|\psi_\delta(u)\|_2\leq\delta\sqrt m$, ensuring all required moments exist
and giving the following bias and variance bounds:
\begin{align}
 \left\|\mathbb E\psi_\delta(u)-\mathbb E_Pu\right\|_2
 &\leq b_\delta+2\pi\delta\sqrt m,\\
 \operatorname{tr}(\operatorname{Cov}(\psi_\delta(u)))
 &\leq\mathbb E\|\psi_\delta(u)\|_2^2\leq m\delta^2.
\end{align}
The bias--variance decomposition and
$\|Mz\|_2\leq\|M\|_{\mathrm{op}}\|z\|_2$ for any $z\in\mathbb R^m$
prove the claimed bound.

These sample-mean results use a fixed threshold and transport. They also
apply conditionally if these are estimated on independent calibration data.
Conditioning on a quantile computed from the same residuals does not preserve
their independence. To relate the two cases, compare a random
$\widehat\delta$ with a fixed $\delta_0$. Lipschitz dependence on the
threshold gives, for any slack parameter $\beta>0$,
\begin{align}
 \mathbb E\|\widehat g_{\widehat\delta}-g_P\|_2^2
 &\leq(1+\beta)\mathbb E\|\widehat g_{\delta_0}-g_P\|_2^2\nonumber\\
 &\quad +(1+\beta^{-1})\|M\|_{\mathrm{op}}^2m
                 \mathbb E|\widehat\delta-\delta_0|^2.
\end{align}
The additional term must be controlled to obtain a sample-dependent threshold
guarantee. Likewise, dependent particles add cross-covariance terms to the
variance of their mean.

\section{Experimental protocols and evaluation details}
\label{app:experimental_details}

\subsection{Solver settings}
\label{app:solver_settings}

\paragraph{Navier--Stokes.}
The nine-setting sweep uses $J=2048$, batch size 64, 80 steps, two updates,
guidance scale 2.0, numerical floor 0.01, and quantile 0.995.
It also sets \texttt{precision\_shrinkage}=1.0,
\texttt{precision\_power}=2.0, observation-noise standard deviation $\sigma_y$,
and \texttt{guidance\_decay\_power}=0.05.

\paragraph{Black-hole imaging.}
The paired launcher uses the official \texttt{nfft} observation backend,
$J=256$, batch size 32, 80 reverse steps, and two updates at a correction.
The numerical floor and quantile are 0.01 and 0.75.
The observation seed is 0 and the inference seed is $100000+\text{case ID}$.

\paragraph{Acoustic FWI.}
The paired ensemble launcher uses $J=64$, batch size 4, four reverse steps,
one update per step, numerical floor 0.01, quantile 0.75, and correction stride 1.
It loads method-specific guidance scales from validation selection.
The observation seed is 0 and the inference seed is $100000+\text{case ID}$.

\subsection{Controlled experiments}
\label{app:controlled_details}

Panels (a) and (b) of Figure~\ref{fig:controlled_mechanisms} use state
dimension 16, observation dimension 64, and $J=256$.
Panel (a) uses spread ratios $1,2,4,8,16,32,64$ without contamination.
Panel (b) fixes the spread ratio to one and the contaminated fraction to 10\%,
with outlier magnitudes $0,5,10,20,40$ times the noise standard deviation.

Panel (c) uses a separate linear experiment with state dimension 12 and
observation dimension 48. It compares analytic and empirical risk gains on
120 generated problems. Population and empirical weighted directions are
rescaled to match the corresponding unweighted linear-update Frobenius norm.
Empirical predictions in this experiment use standard deviations with
denominator $J$.

\subsection{Metrics and visualization conventions}
\label{app:metrics}

For a displayed reconstruction $\widehat x$, relative error is
$\|\widehat x-x^\star\|_2/\|x^\star\|_2$.
Absolute-error images display $|\widehat x-x^\star|$.
NS qualitative scores use the displayed particle mean.

Paired FWI reconstruction metrics are evaluated on the ensemble mean.
The saved seismic misfit is $\sqrt{\tfrac12\|G(\widehat x)-y\|_2^2}$,
with case metrics then aggregated.
The NS and paired FWI records use population standard deviations.

Black-hole qualitative panels use the existing translation-alignment grid
and the saved particle with the highest aligned PSNR against the reference
image. Both ensemble methods use the same rule.
Their blurred PSNR column corresponds to the saved evaluator field with
blur factor $f=20$. EnKG and \method{} share observations; other qualitative
baselines retain native observation realizations for the same target and
observation-time setting.

In Figure~\ref{fig:navier_qualitative}, NS reconstructions share a symmetric
vorticity scale. Absolute-error colors use a common square-root mapping
with ticks in the original error units.
Black-hole panels share the range $[0,1]$.

\clearpage
\section{Additional qualitative results}
\label{app:qualitative}

\subsection{Navier--Stokes inversion}
\label{app:qualitative_ns}

\begin{figure}[htbp]
    \centering
    \includegraphics[width=\linewidth]{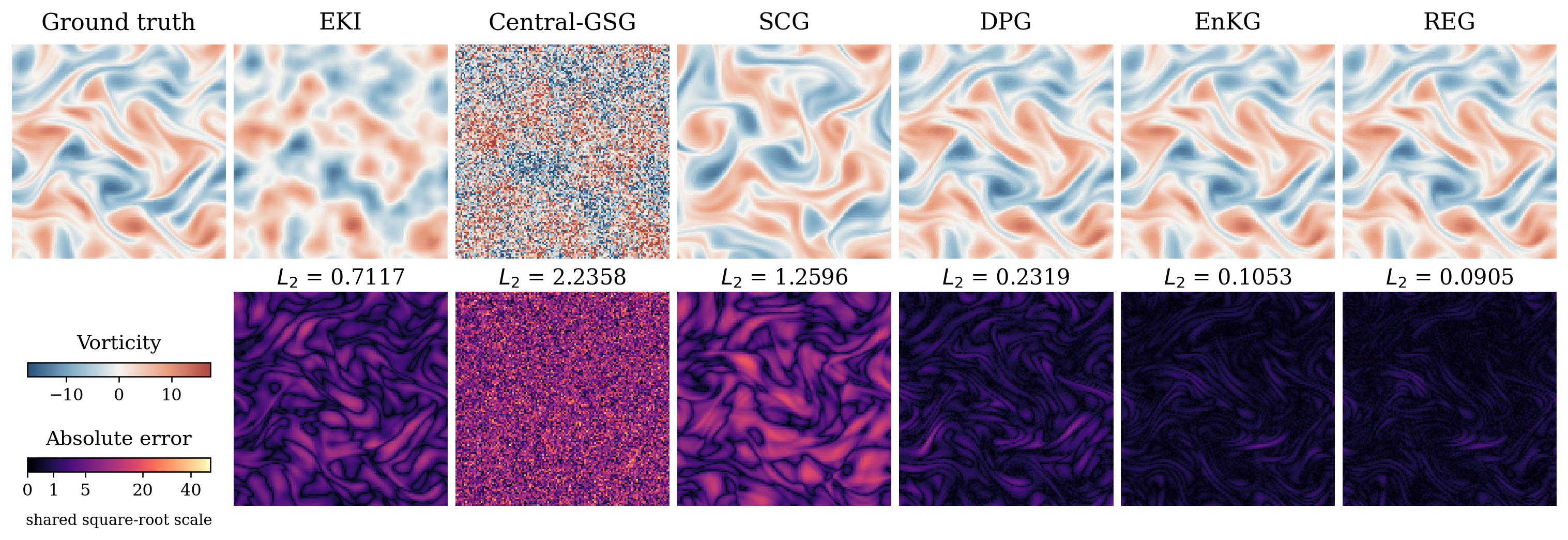}
    \caption{NS reconstructions and absolute errors at $\times2$,
    $\sigma_y=0$. EnKG and \method{} use $J=2048$; numbers are relative
    $\ell_2$ errors of the displayed reconstructions.}
    \label{fig:app_ns_multimethod}
\end{figure}

Figure~\ref{fig:app_ns_multimethod} extends the multi-method comparison to
another target under the original matched configuration.
\method{} preserves the vortex structure and attains the lowest
reconstruction error among the displayed methods.

\begin{figure}[htbp]
    \centering
    \includegraphics[width=0.80\linewidth]{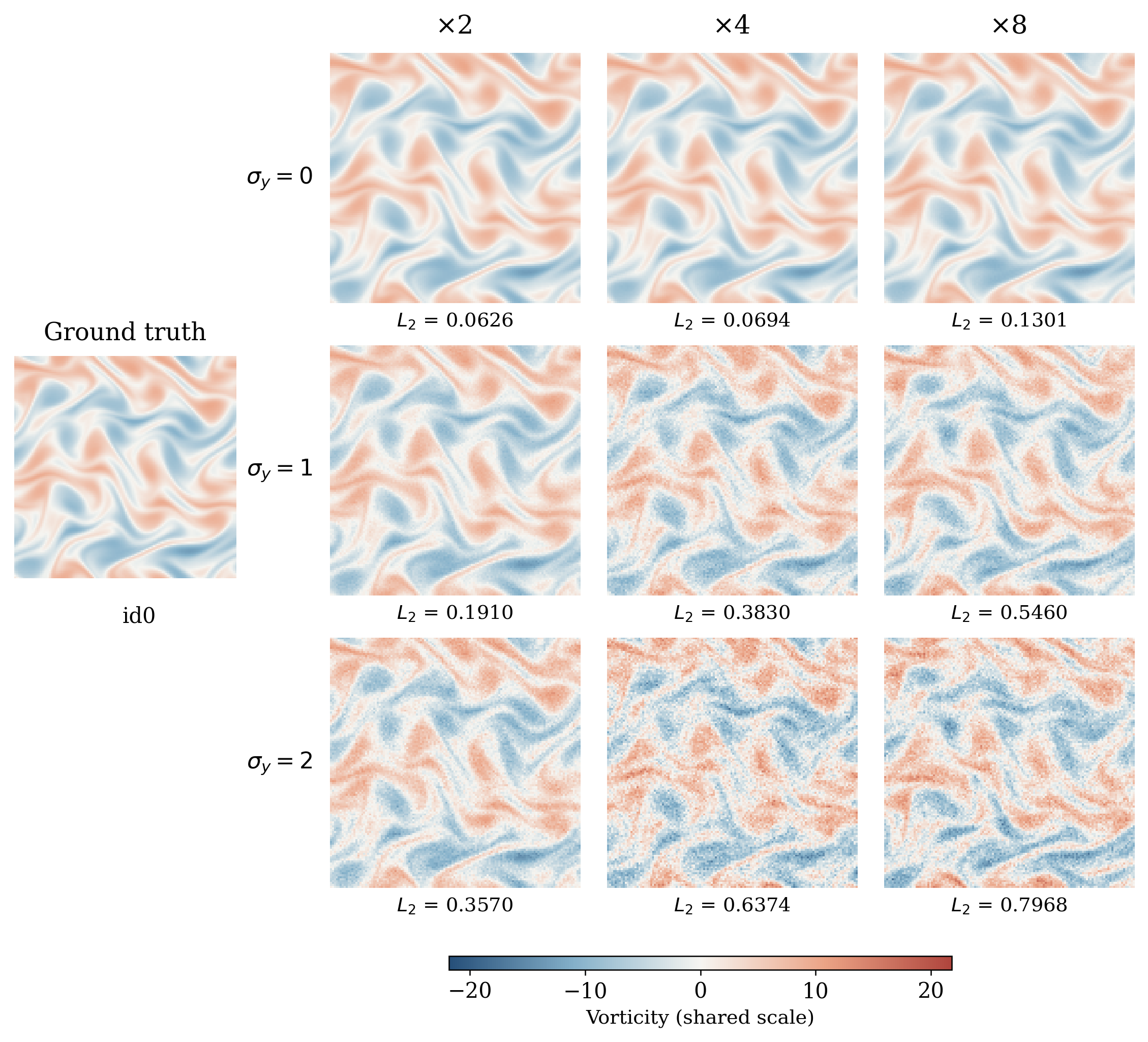}
    \caption{\method{} reconstructions of NS id0 across downsampling factors
    $\times2/4/8$ and noise levels $\sigma_y=0/1/2$, with $J=2048$.
    All panels share a vorticity scale.}
    \label{fig:app_ns_nine_settings}
\end{figure}

Figure~\ref{fig:app_ns_nine_settings} holds the target fixed across the nine
settings using the noise-aware configuration in Appendix~\ref{app:solver_settings}.
Dense, noiseless observations give the most accurate recovery, while stronger noise
and coarser sampling progressively obscure finer structures.

\clearpage
\subsection{Black-hole imaging}
\label{app:qualitative_blackhole}

\begin{figure}[htbp]
    \centering
    \includegraphics[width=0.76\linewidth]{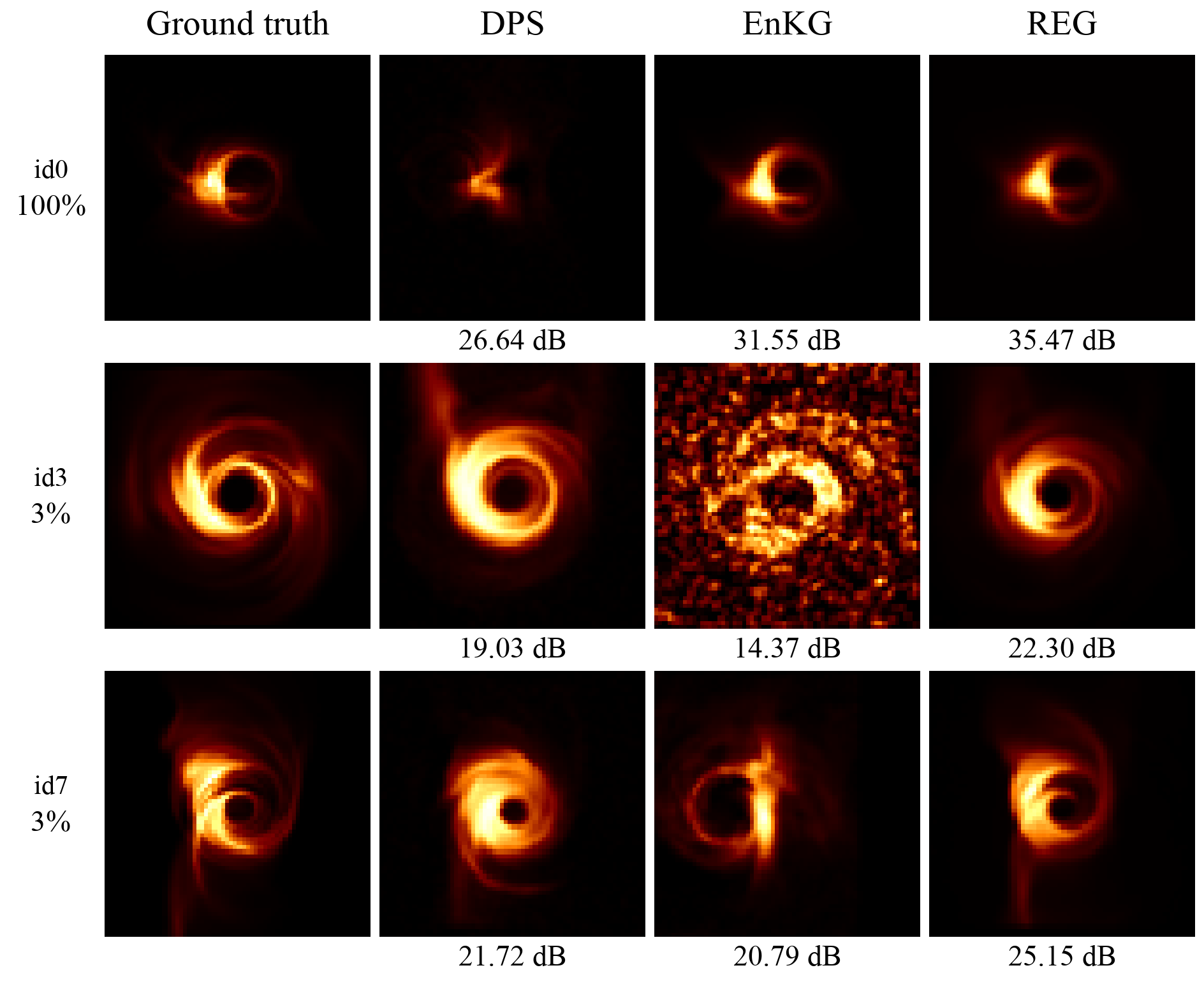}
    \caption{Selected black-hole reconstructions: id0 at $100\%$, id3 at $3\%$,
    and id7 at $3\%$ observation time. Numbers are aligned PSNR (dB);
    evaluation follows Appendix~\ref{app:metrics}.}
    \label{fig:app_bh_selected}
\end{figure}

Figure~\ref{fig:app_bh_selected} compares three additional targets.
DPS loses part of the ring or distorts its shape, while EnKG shows structural
distortion or strong granular artifacts. \method{} better preserves the
ring and brightness distribution in these examples.

\begin{figure}[htbp]
    \centering
    \includegraphics[width=0.96\linewidth]{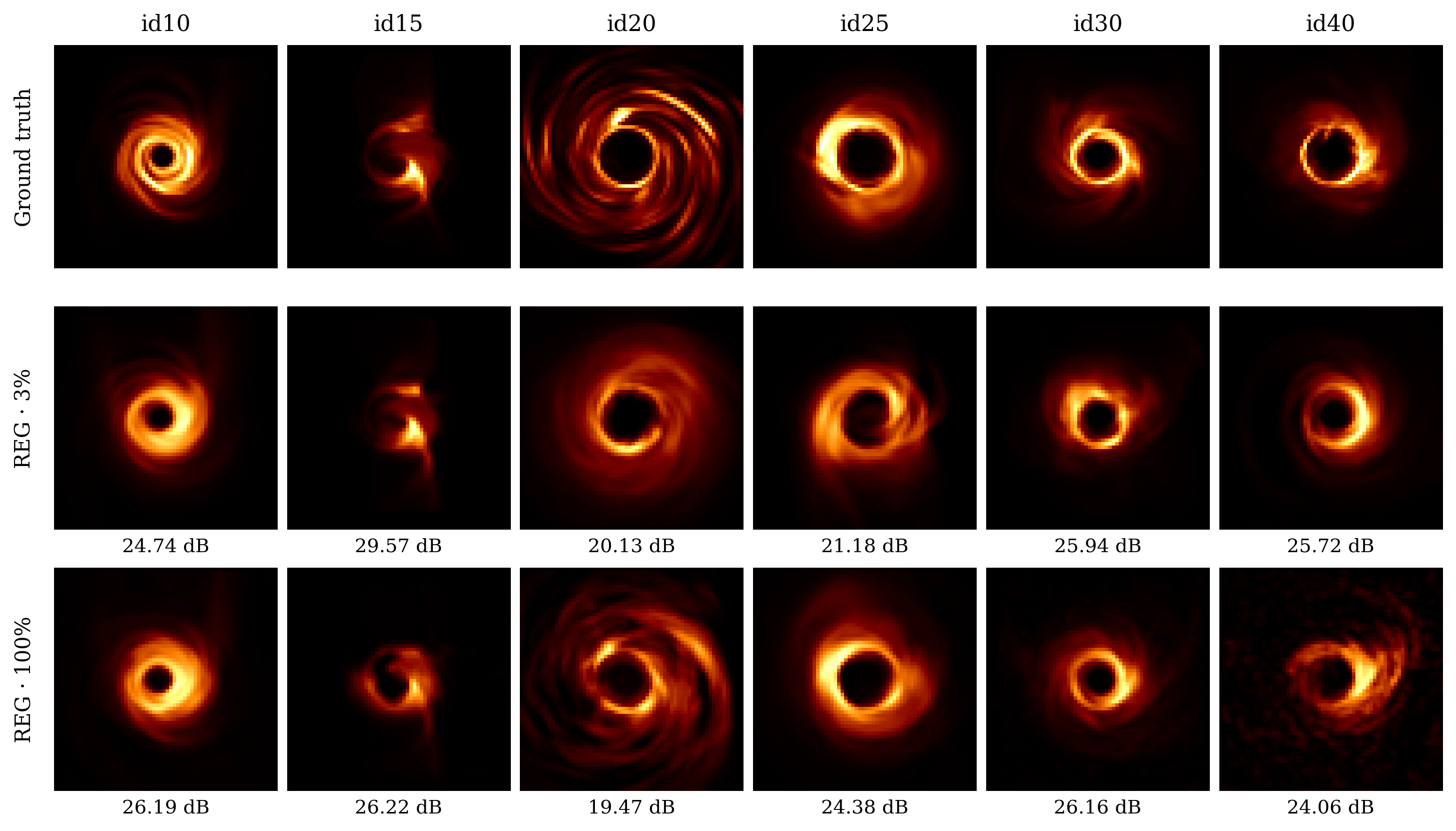}
    \caption{Additional black-hole targets and \method{} reconstructions at
    $3\%$ and $100\%$ observation time, with $J=256$.
    Numbers are aligned PSNR (dB).}
    \label{fig:app_bh_more_targets}
\end{figure}

Figure~\ref{fig:app_bh_more_targets} presents six further targets at two
observation times. \method{} recovers the main ring and bright regions
across different morphologies, while the intricate outer spiral patterns
are less fully resolved.

\clearpage
\subsection{Acoustic full-waveform inversion}
\label{app:qualitative_fwi}

\begin{figure}[htbp]
    \centering
    \includegraphics[width=\linewidth]{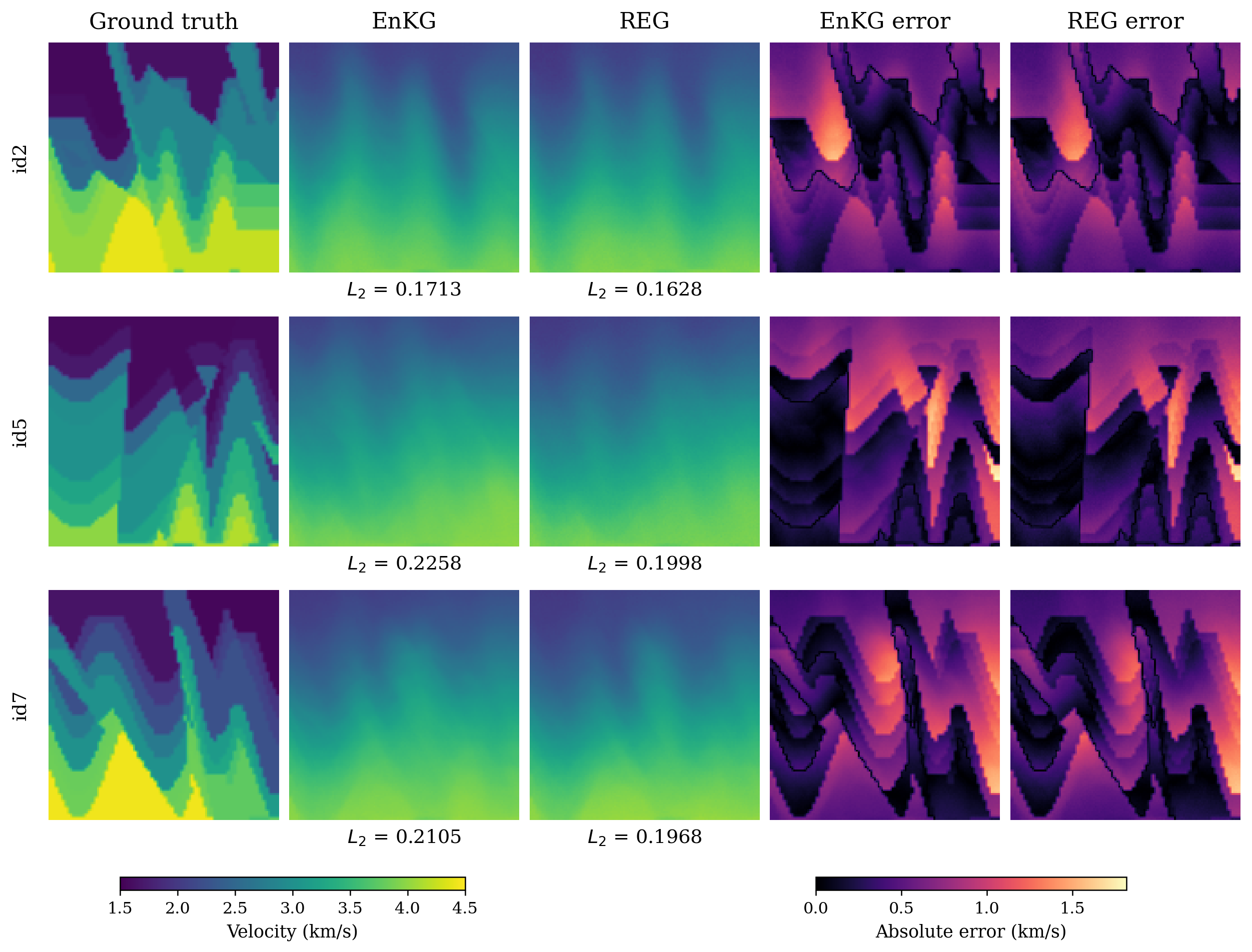}
    \caption{FWI reconstructions and absolute errors for id2, id5, and id7,
    with $J=64$. Velocity and error panels use shared color scales;
    numbers are relative $\ell_2$ errors.}
    \label{fig:app_fwi_paired}
\end{figure}

Figure~\ref{fig:app_fwi_paired} compares EnKG and \method{} on three velocity
fields. \method{} reduces reconstruction error in each case and better
matches the broad velocity variations. Fine interfaces remain smoothed
in both methods.

\end{document}